\documentclass{article} 
\usepackage{iclr2027_conference,times}

\usepackage{amsmath,amsfonts,bm}

\def\eqref#1{equation~\ref{#1}}

\def\1{\bm{1}}

\DeclareMathAlphabet{\mathsfit}{\encodingdefault}{\sfdefault}{m}{sl}
\SetMathAlphabet{\mathsfit}{bold}{\encodingdefault}{\sfdefault}{bx}{n}

\usepackage{hyperref}
\usepackage{url}

\usepackage{amsmath}
\usepackage{hyperref}
\usepackage{url}
\usepackage{pifont} 
\usepackage{graphicx}
\usepackage{booktabs}
\usepackage{xspace}
\usepackage{xcolor}
\usepackage{colortbl} 

\definecolor{dpos}{HTML}{2E7D32}  
\definecolor{dneg}{HTML}{C62828}  
\definecolor{dzero}{HTML}{8A8A8A} 
\definecolor{bestcell}{HTML}{D8E9F5} 
\definecolor{secondcell}{HTML}{EDF4FA} 
\definecolor{bandA}{HTML}{E7F0E4} 
\definecolor{bandB}{HTML}{FDF0DC} 
\definecolor{bandC}{HTML}{EEE8F4} 
\definecolor{bandD}{HTML}{ECECEC} 
\definecolor{bandE}{HTML}{FBE4E4} 
\definecolor{bandF}{HTML}{E0F1F0} 
\definecolor{bandG}{HTML}{FBF4D9} 
\definecolor{bandH}{HTML}{EFE6DC} 
\definecolor{bandI}{HTML}{E4F2EA} 
\definecolor{bandJ}{HTML}{F3E6F0} 
\usepackage{enumitem}
\definecolor{figblue}{HTML}{3D6A8F}
\definecolor{figorange}{HTML}{9E6132}
\definecolor{figgreen}{HTML}{3E7D5A}
\definecolor{figred}{HTML}{A94F42}
\usepackage[capitalise]{cleveref} 

\newcommand{\MODEL}{TimeEvo\xspace}
\newcommand{\codeurl}{https://github.com/Muyiiiii/TimeEvo}
\newcommand{\codelink}{\url{\codeurl}}

\title{\MODEL: Failure-Driven Self-Evolution of a Time Series Agent}

\author{%
Jie Yang$^{1,2}$, Yan Zheng$^{2}$, Jiarui Sun$^{2}$, Xiran Fan$^{2}$, Junpeng Wang$^{2}$, Liang Wang$^{2}$, \\
\bfseries Zelin Xu$^{2,3}$, Qinghua Liu$^{2,4}$, Zhengyu Fang$^{2,5}$, Yiwei Cai$^{2}$, Philip S. Yu$^{1,\dagger}$ \\[4pt]
\mdseries
$^{1}$University of Illinois at Chicago \quad
$^{2}$Visa Research \quad
$^{3}$University of Florida \\
$^{4}$The Ohio State University \quad
$^{5}$Case Western Reserve University \\[2pt]
\texttt{jyang265@uic.edu}
}

\iclrfinalcopy 
\begin{document}

\maketitle
\lhead{Preprint.}
{\renewcommand{\thefootnote}{\ensuremath{\dagger}}\footnotetext{Corresponding author.}}

\begin{abstract}
Time series agents answer analytical questions by calling external tools, and which tools they carry is decided by people before the agent runs.
However, we identify two failures in this setup.
\textbf{Human--Agent Tool Misalignment}: a library of 21 expert-curated tools helps on some tasks and hurts on others, dropping anomaly accuracy under every backbone we test.
\textbf{Silent Harm}: one round of generic self-revision changes 147 answers and breaks 56 of them, while the final score moves by less than a point.
Both follow from the same gap: whether a tool helps is decided question by question at runtime, while tools are supplied in advance and judged by a single average.
To address this, we propose \MODEL, which clusters an agent's diagnosed failures into capability gaps, plans a measurement for each, synthesizes evidence-only tools that fill them, and admits the candidate library only through a paired admission gate.
Experiments on ten time series QA tasks and three backbones show that \MODEL, starting from an empty library, improves accuracy on every task and every backbone, and that a library grown on a cheap model still gains when it is installed into stronger ones.
Code is available at \mbox{\codelink}.
\end{abstract}

\section{Introduction}
\label{sec:intro}

Time series analysis~\citep{hulandscape,beqari2024pilot} is a fundamental problem with broad impact in real-world applications such as finance~\citep{yang2025grad,cheng2025neighbor}, traffic~\citep{hu2026bridging}, and healthcare~\citep{yang2025revisiting}, and spans a wide range of tasks, including forecasting~\citep{yang2026observations,wang2025fredf}, imputation~\citep{yang2026glocal}, and time series understanding~\citep{xie2024chatts}.
To address this diverse task landscape, model-based approaches follow two broad designs.
Dedicated models~\citep{liu2024itransformer, zeng2023transformers} provide efficient and competitive performance on well-defined objectives, whereas multimodal language models~\citep{jin2024time,xie2024chatts} treat time series as an additional modality for natural-language interaction and open-ended questions~\citep{kong2025time}.
Despite their different interfaces, both acquire analytical capabilities through task-oriented training and encode them in model parameters.
Therefore, supporting a new task or capability typically requires corresponding data construction, training, or tuning~\citep{dong2026agentvaluebenchcomprehensivebenchmarkevaluating,kong2026timesage}.

Large language model~(LLM) agents~\citep{yao2022react,zhu2026rsiagentautonomousexplorationrecursive} offer an alternative: rather than internalizing every time series operation in model parameters~\citep{NEURIPS2025_34aae5b7}, they interpret user requests, plan analyses, and invoke external analytical tools~\citep{wu2026timeart}.
Across tasks ranging from forecasting~\citep{weng2026temporalbench} to open-ended time series understanding~\citep{kong2025time,yu2026tsrouter}, they synthesize tool outputs and intermediate evidence into coherent natural-language responses~\citep{ye2024ts, zhao2025timeseriesscientist}.
This design implements many domain-specific analytical operations as reusable external modules that can be inspected, recombined, and modified without retraining the underlying model~\citep{wu2026timeart}.
Building on this modular design, recent agent frameworks further refine how agents use these external modules by distilling execution experience into reusable guidance~\citep{liu2026timeclaw, ye2024ts}.

However, task-specific development persists: human effort does not disappear but shifts from model training~\citep{fan2026mosaic,fan2026trace,fang2026novel} to configuring prompts, workflows, and tools for tasks anticipated before deployment~\citep{hao2026poise}.
It remains unclear whether supplying an agent with more tools makes it better at the tasks those tools were written for, and whether the agent's own revisions can be trusted to improve it~\citep{miao2025recode}.
To investigate, we analyze the errors and repair attempts of a frozen-LLM time series agent~(TSAgent) across time series QA tasks as shown in~\cref{fig:intro}.
Our analysis reveals two key phenomena:

\begin{figure}[t]
  \centering
  \includegraphics[width=\textwidth]{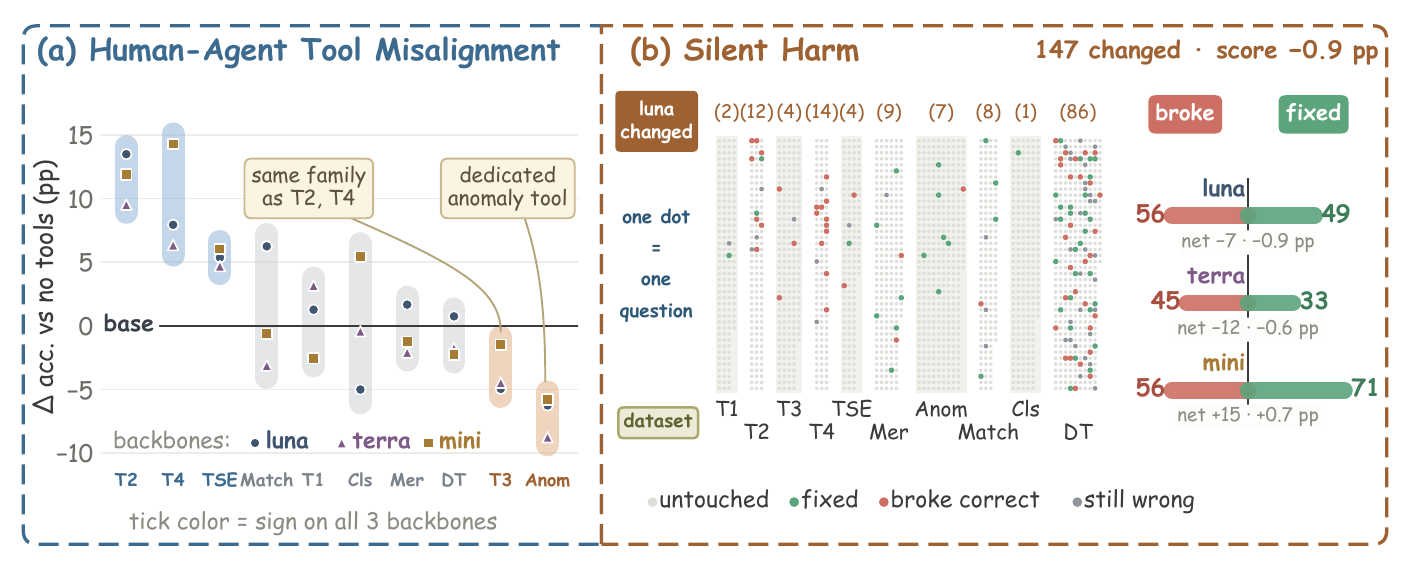}
  \caption{
    \textbf{Two failure phenomena in TSAgent.}
    We compare the same frozen agent with and without 21 expert-curated tools~\textbf{(a)}, and before and after one generic self-refinement pass~\textbf{(b)}, pairing every question across ten QA tasks and three backbone LLMs.
    Improved and degraded tasks are shaded \textcolor{figblue}{\textbf{blue}} and \textcolor{figorange}{\textbf{orange}} in~(a), and fixed and broken answers are shown in \textcolor{figgreen}{\textbf{green}} and \textcolor{figred}{\textbf{red}} in~(b).
  }
  \label{fig:intro}
\end{figure}

\begin{itemize}[leftmargin=*]
    \item[\ding{182}]\label{phe:1}
    \textbf{\textcolor{figblue}{Human--Agent Tool Misalignment}: \emph{what humans prefer to supply is not what the agent benefits from.}}
Equipping the same frozen agent with a library of 21 expert-curated analysis tools of TimeART~\citep{wu2026timeart} makes it better on some tasks and worse on others~(\cref{fig:intro}a).
The largest gains and the largest losses both recur in the same direction under all three backbone LLMs.
The losses are not a coverage problem: the library includes a dedicated anomaly-detection tool, yet the anomaly task drops by 5.8 to 8.8 points under every backbone.
T3 shares the same task family with the improving T2 and T4~\citep{weng2026temporalbench}, yet it is degraded under all three backbones.
The misalignment is structural: the supply is fixed per task before deployment, but whether a tool helps is decided question by question at runtime.

    \item[\ding{183}]\label{phe:2}
    \textbf{\textcolor{figorange}{Silent Harm}: \emph{self-revision breaks answers that the final score never shows.}}
A single pass of the Self-Refine~\citep{madaan2023self} changes 147 answers across the ten tasks: 49 are fixed, 56 previously correct answers break, and the remaining 42 swap one wrong answer for another.
Yet the final score moves by less than one point~(\cref{fig:intro}b).
The pattern is not specific to one model: on two further backbones the pass touches 97 and 163 answers, breaks 45 and 56 of them, and again never moves the final score by a full point.
On T4 alone, the pass on luna changes 14 answers, fixes 0, and breaks 11.
The silence is arithmetic: inside a single average, the harm silently erases the fixes, and the final score never moves.
\end{itemize}

Together, these findings show that the tools agents are given do not match what they need at runtime, while unexamined self-revision silently erases much of what it fixes.
This leads to a central question: \emph{\textbf{Can a time series agent supply its own missing tools from diagnosed failures while measurably protecting what it already answers correctly from silent harm?}}

To answer this question, we propose \textbf{\MODEL}, a self-evolving agent for time series analysis.
The key idea is to let the agent discover the tools it actually needs from the failures it makes, and to trust no self-made update until it proves that it helps more than it harms.
Specifically, \MODEL clusters the agent's errors into failure buckets, plans a measurement contract for each, and synthesizes evidence-only tools to fulfill them.
The whole candidate library must then pass two levels of validation, and a failing library is pruned and retested once.
Ultimately, the agent evolves its own task-specific tool library, in which every tool has earned its place without harming silently.

Our main contributions are summarized as follows:
\begin{itemize}[leftmargin=*]
    \item We identify two phenomena in time series agents, \textbf{Human--Agent Tool Misalignment} and \textbf{Silent Harm}: what humans prefer to supply is not what the agent benefits from, and self-revision silently erases much of what it fixes.
    \item We propose \textbf{\MODEL}, a self-evolving agent that discovers the tools it needs from the failures it makes, and admits every self-made update only after two levels of validation.
    \item Extensive experiments on ten time series QA benchmarks show that \MODEL improves its base agent on every task, with pooled gains from $+1.5$ to $+14.7$ points.
    Neither expert-curated tools nor generic self-refinement reproduces this.
\end{itemize}
\section{Related Work}
\label{sec:related_work}

\subsection{Time Series Analysis}
Time series analysis has long been dominated by specialized models built for individual tasks such as forecasting~\citep{yang2026observations}, classification~\citep{dempster2020rocket}, and anomaly detection~\citep{liu2024elephant}.
Pretrained foundation models~(e.g., Chronos~\citep{ansari2025chronos}, TimesFM~\citep{das2023decoder}, and Moirai~\citep{liu2025moirai}) extend this line by transferring forecasting capability across domains.
Furthermore, the emergence of LLMs has enabled more flexible and powerful approaches to time series analysis.
Multimodal language models such as ChatTS~\citep{xie2024chatts} and TimeLLM~\citep{jin2024time} take time series as an additional input modality and answer open-ended questions end to end.
Yet facing a new task, these models still depend on collecting task-specific data and training or tuning the model.
In contrast, agent-based systems~(e.g., TS-Reasoner~\citep{ye2024ts}, TS-Agent~\citep{liu2025ts}, TimeCopilot~\citep{garza2025timecopilot}) offer a training-free alternative that moves analytical capabilities outside the model: the agent plans over external tools and synthesizes their outputs into natural-language answers.
Training-free, however, does not mean effort-free: human effort merely shifts from model training to designing prompts, workflows, and tools before deployment.
To reduce this reliance on human design, we propose \MODEL, which treats diagnosed failures as the supervision signal for evolving the capability library itself.

\subsection{Self-Evolving Agents}
Self-evolution has been extensively explored for general LLM agents.
One line, exemplified by Reflexion~\citep{shinn2023reflexion} and EvolveR~\citep{wu2025evolver}, distills experience from past trajectories into reusable lessons or principles.
Another line maintains growing skill libraries such as Voyager~\citep{wang2023voyager}, and recent work like SkillOpt~\citep{yang2026skillopt} and CoEvoSkills~\citep{zhang2026coevoskills} further tests learned skills before deployment with held-out scores or per-task checks.
Within time series, TimeClaw distills usage experience over a fixed tool library~\citep{liu2026timeclaw}.
However, these frameworks admit an update through the final score or a per-task check at best, and some skip validation entirely.
The admission never hinges on how many previously correct answers the update breaks.
Inside a single final score, such breaks cancel against the fixes, so an update can cause real harm while the score barely moves, exactly the Silent Harm identified in our analysis.
This motivates \MODEL, which evolves the tool library from diagnosed failures and admits every update only after two levels of validation that weigh what it fixes against what it breaks.
\section{Method}
\label{sec:method}

\begin{figure}[t]
  \centering
  \includegraphics[width=\textwidth]{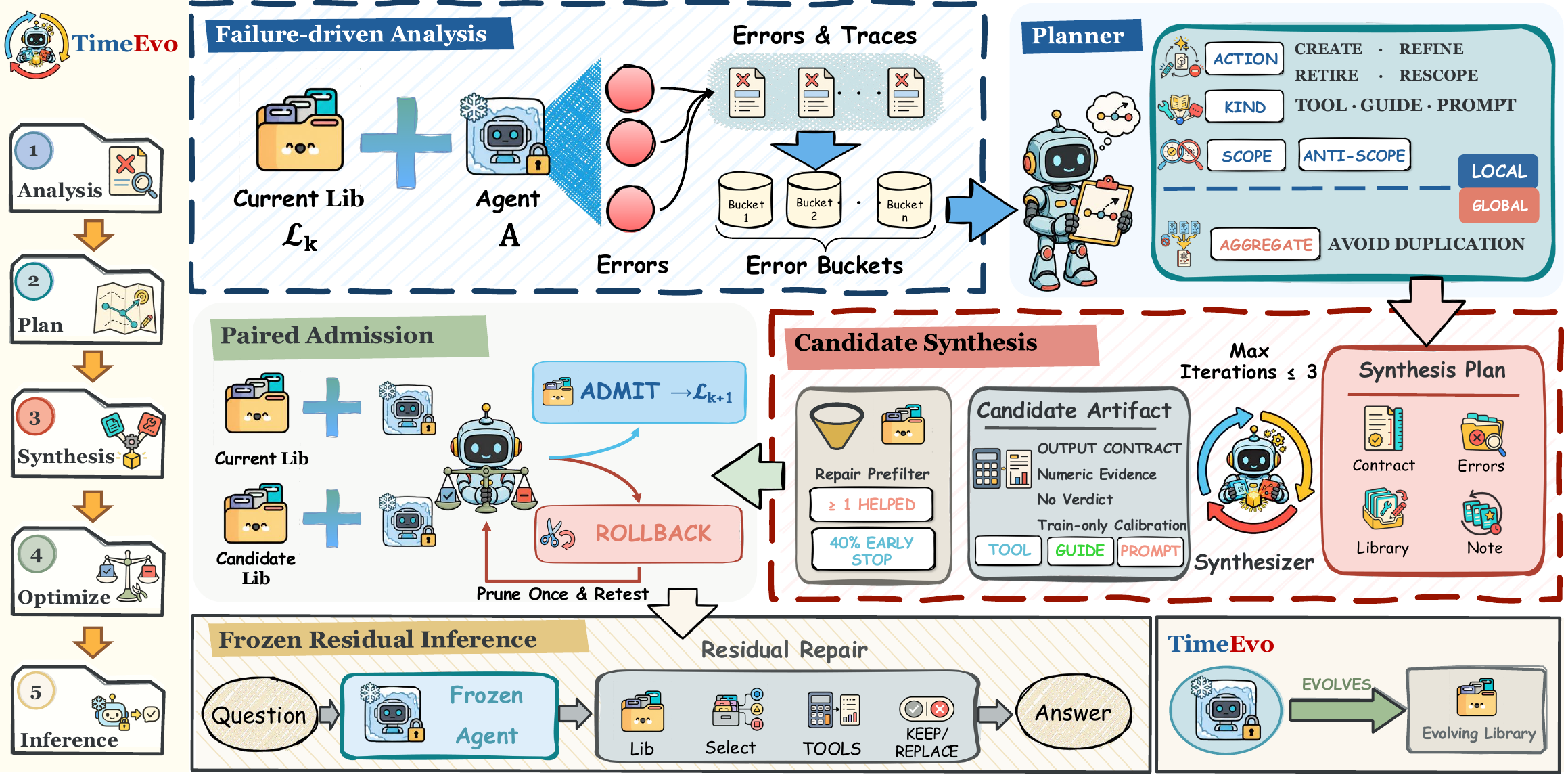}
  \caption{
    \textbf{Overview of \MODEL.}
    Each round turns the agent's own failures into new tools: errors are clustered into failure buckets, each bucket is planned into a measurement contract, and contracts are synthesized into evidence-only candidate tools.
    The candidate library is admitted only after two levels of validation, a per-tool prefilter and a whole-library paired gate. A failing library is pruned and retested once.
    Admitted tools join the agent's library, while rejected rounds leave the agent untouched.
    At inference, the frozen agent answers first, and admitted tools review only in-scope questions, keeping or replacing the answer.
  }
  \label{fig:method}
\end{figure}

\subsection{Problem Definition}
\label{sec:problem}

Each time series question is a triple $x = (q, \mathbf{S}, \mathcal{O})$: a natural-language question $q$ that may carry event context, an observed input $\mathbf{S}$ collecting one or two primary series together with any named covariate or candidate series the question refers to, and a finite option set $\mathcal{O}$.
An agent is a frozen LLM equipped with a library $\mathcal{L}$ of executable analysis tools, and it answers by combining tool evidence with its own reasoning to select one option from $\mathcal{O}$.
We write $\mathcal{A}_{\mathcal{L}}$ for the agent carrying $\mathcal{L}$, and the base agent starts from an empty library $\mathcal{L}_0$~(a pre-installed forecasting root is kept only as an ablation).
The gold answer is determined by the observed input alone, so whether the agent answers a question correctly can be checked automatically.
A final score alone cannot judge a library update $\mathcal{L} \rightarrow \mathcal{L}'$, which is the Silent Harm of~\cref{fig:intro}b.
We therefore track what an update fixes and what it breaks on a question set $D$:
\begin{equation}
\begin{aligned}
\mathrm{helped}(D) &= \{x \in D : v(x) = 0,\ v'(x) = 1\},\\
\mathrm{harmed}(D) &= \{x \in D : v(x) = 1,\ v'(x) = 0\},
\end{aligned}
\label{eq:paired}
\end{equation}
where $v$ and $v'$ denote correctness under $\mathcal{L}$ and $\mathcal{L}'$ on the same questions.
Given train, validation, and test splits $D_{\mathrm{train}}, D_{\mathrm{val}}, D_{\mathrm{test}}$, the goal is to evolve $\mathcal{L}$ from the agent's failures on $D_{\mathrm{train}}$ so that helped outnumbers harmed on unseen questions, while $D_{\mathrm{test}}$ is never used for any decision.

\subsection{Overview and Frozen Residual Inference}
\label{sec:overview}

\MODEL runs in rounds, and each round turns the agent's current failures into new tools~(\cref{fig:method}).
A round has four stages: diagnose the failures on the training split into measurement contracts, synthesize evidence-only tools for the contracts, screen each tool with a cheap prefilter, and submit the surviving library as a whole to a paired admission gate.
After $r$ accepted rounds the deployed library is a chain $\mathcal{L}_r = (\mathcal{L}_0, \mathcal{T}_1, \ldots, \mathcal{T}_r)$, where $\mathcal{T}_k$ is the tool set admitted in round $k$, and a question is threaded through the stages:
\begin{equation}
\mathcal{A}_{\mathcal{L}_k}(x) =
\begin{cases}
\rho_k\big(x,\ E_k(x),\ \mathcal{A}_{\mathcal{L}_{k-1}}(x)\big) & \text{if } x \in \mathrm{scope}(\mathcal{T}_k),\\[2pt]
\mathcal{A}_{\mathcal{L}_{k-1}}(x) & \text{otherwise},
\end{cases}
\label{eq:chain}
\end{equation}
where $\mathrm{scope}(\mathcal{T}_k)$ matches the stage's declared scope fields~(task types, evidence types, single- or dual-series input, and an anti-scope), $E_k(x)$ is the numeric evidence the stage's tools compute on $x$, and $\rho_k$ is the same agent reviewing the incoming answer with that evidence.
At answer time, as the \emph{Frozen Residual Inference} strip of~\cref{fig:method} shows, the agent consults a structured scope catalog and inspects at most three shortlisted tools per question.
The review is conservative by construction: the incoming answer is the default, and a replacement is honored only when a stage tool actually ran on the question.
If a tool errors, is never invoked, or the review does not explicitly decide to replace, the question falls back to the incoming answer.
Out-of-scope questions are therefore never degraded. 
In-scope harm remains possible, and the admission gate measures it alongside repair and screens candidate libraries for positive net improvement on validation data.

\subsection{Failure-Driven Analysis and Planning}
\label{sec:diagnose}

As shown in the \emph{Failure-driven Analysis} panel of~\cref{fig:method}, round $r$ first replays the current library on the training split, collects the questions it answers wrong, and partitions them into failure buckets:
\begin{equation}
\mathcal{F}_r = \{x \in D_{\mathrm{train}} : v_{r-1}(x) = 0\},
\qquad
\mathcal{F}_r = \mathcal{B}_1 \cup \cdots \cup \mathcal{B}_J,
\label{eq:buckets}
\end{equation}
where $v_{r-1}$ denotes correctness under the current library $\mathcal{L}_{r-1}$ and the buckets are disjoint.
The agent builds this partition in two steps: it first defines a small set of failure categories from a sample of the errors, then assigns every error to exactly one category.
Errors on which a tool fired and errors that no tool reached are clustered separately, because the former indicate a defective or misused tool while the latter indicate a coverage gap.
The definition step never sees task metadata, so buckets group errors by the missing capability rather than by question type. Oversized buckets are split once, and buckets with fewer than two errors are set aside.

Each bucket $\mathcal{B}_j$ is then diagnosed by a local planner~(the \emph{Planner} panel in~\cref{fig:method}), which must return a \emph{measurement contract} $c_j$: the diagnosed root cause, the required measurement, the inputs it needs, a success criterion, and a scope with an anti-scope of applicability.
The contract also declares one library action, typed by the diagnosed failure mode: \textsc{\textbf{create}} a new tool when evidence is missing, \textsc{\textbf{refine}} an existing tool with usage guidance when its evidence is ignored or misread, \textsc{\textbf{rescope}} a tool that fires on the wrong questions, and \textsc{\textbf{retire}} a tool that should leave the library.
Because buckets are planned in parallel, a global planner coordinates the proposals afterwards: duplicated measurements are merged and overlapping scopes are narrowed, while a merge of incompatible contracts is rejected mechanically and falls back to the independent per-bucket plans.
Planning is thus committed to measurements rather than answers: what a bucket receives is a contract for the evidence its failures lack, not a rule for how to answer them.

\subsection{Candidate Synthesis and Prefilter}
\label{sec:synth}

As shown in the \emph{Candidate Synthesis} panel of~\cref{fig:method}, each coordinated contract $c_j$ receives up to three sequential synthesis attempts:
\begin{equation}
t_j^{(i)} = \mathrm{Synth}\big(c_j,\ f_j^{(i-1)}\big),
\qquad
t_j^{(i)} : (\mathbf{S}, q) \mapsto \mathbf{e} \in \mathbb{R}^{d_j},
\qquad i \le 3,
\label{eq:synth}
\end{equation}
where $f_j^{(i-1)}$ is the structured feedback of the previous failed attempt~($f_j^{(0)} = \emptyset$), and the delivered tool $t_j$ is the first passing attempt, one deterministic Python function.
A tool is evidence-only: it never outputs a verdict, an option string, or a dataset-specific constant.
It may read a covariate only when the question mentions it, and when its input is unavailable it abstains rather than fabricates a value.
A sandbox that permits only a fixed allowlist of operations enforces determinism and isolation, and a candidate that fails any of these checks is discarded.
A tool is not limited to plain arithmetic on the series: it may call the raw Chronos-2 model~\citep{ansari2025chronos} as a forecasting primitive, and when its contract requests a decision boundary it may carry a \emph{train-only calibration}, a shallow decision tree fitted on training rows alone and compiled into its source:
\begin{equation}
g_j = \mathrm{Fit}\big(\{(\,t_j(x),\ y(x)\,) : x \in D_{\mathrm{train}}\}\big),
\qquad
t_j \leftarrow t_j \oplus g_j \ \ \text{iff}\ \ \mathrm{BalAcc}(g_j) \ge 0.55,
\label{eq:calib}
\end{equation}
where $y(x)$ is the gold answer, $\oplus$ denotes compiling the fitted tree into the tool's source code, and $\mathrm{BalAcc}$ is balanced accuracy on an internal train-side split with at least 16 samples per class.
Thresholds are therefore measured on training rows rather than invented by the agent.

The first level of validation is a per-tool prefilter, run on at most 32 held-out errors $\mathcal{H}_j \subset \mathcal{B}_j$ of the tool's own bucket.
A candidate survives only if it delivers at least one \emph{attributable} repair: a question fixed with the tool actually attached, actually executed, and visible in the execution record.
It must also break few previously correct answers and rarely fail to execute.
Two consecutive attempts without an attributable repair send the bucket back for re-diagnosis, which may change the artifact kind, for example from a new tool to usage guidance.
The prefilter only discards clearly useless candidates: every statistical decision is reserved for the admission gate, the only decision whose mistakes would reach deployment.

\subsection{Paired Admission}
\label{sec:gate}

The second level of validation~(the \emph{Paired Admission} panel in~\cref{fig:method}) decides admission for the round's surviving tools $\mathcal{T}_r$ as a whole rather than tool by tool, because tools that look harmless in isolation can interfere once deployed together.
The current library and the candidate library answer the full validation split in one paired run, and only questions that both sides complete are compared.
Applying~\cref{eq:paired} to $D_{\mathrm{val}}$ gives $h = |\mathrm{helped}(D_{\mathrm{val}})|$ and $m = |\mathrm{harmed}(D_{\mathrm{val}})|$. Over the $n$ paired questions this yields the gate statistic:
\begin{equation}
\delta = \frac{h - m}{n},
\qquad
\delta_{\mathrm{lb}} = \delta - Z \cdot \frac{\sqrt{\max(h + m,\, 1)}}{n},
\qquad Z = 1.96,
\label{eq:strict}
\end{equation}
and the library passes strictly when $h > 0$ and $\delta_{\mathrm{lb}} > 0$. A clean 3-versus-0 round on a small split still fails, because its lower bound stays negative.
A single-round run has no later rounds in which to accumulate evidence, so a supplementary rule admits a library that satisfies:
\begin{equation}
\delta > 0,
\qquad
h - m \ \ge\ \max\big(3,\ \lceil 0.02\,n \rceil\big),
\qquad
\frac{m}{n} \ \le\ 0.10,
\label{eq:safe}
\end{equation}
that is, a positive paired delta, a net repair of at least three questions or two percent of the split, and an absolute harm rate within ten percent.

When the gate rejects the library, \MODEL attempts one repair before giving up.
The validation records show which tool fixed or broke which question, so every update $u$ receives its own counts $h_u$ and $m_u$, and an update is removed only when it is clearly harmful:
\begin{equation}
\mathcal{P} = \big\{\, u \in \mathcal{T}_r \ :\ m_u > 0 \ \wedge\ \big(h_u = 0 \ \vee\ m_u - h_u \ge 3\big) \,\big\}.
\label{eq:prune}
\end{equation}
A tool is left alone whenever the evidence against it is thin: one that changed no answers may simply have met no matching questions, and one whose counts are small and mixed cannot be told apart from noise.
The pruned library is retested in exactly one more full paired run and commits only if it strictly improves the lower bound.
There is no second prune, because repeated retesting would slowly fit the library to the validation split.
Whether a library passes at once or only after the prune, it then reconciles its scope fields against how its tools behaved, and because that edit changes the library, it passes the same paired gate one final time before joining the chain.
A rejected round changes nothing.

\section{Experiments}
\label{sec:experiments}

\begin{table}[t]
\scriptsize
\caption{Main comparison on the ten tasks across three backbones. Each cell shows test accuracy with the within-run paired delta~(pp) in parentheses, where \colorbox{bestcell}{\textbf{bold}} marks the best method per column within a block and \colorbox{secondcell}{\underline{underline}} the second best. The last block installs the library that GPT-5.6-luna evolved into three stronger models, and its \colorbox{bestcell}{shaded} rows carry that library and are not ranked.}
\label{tab:main}
\setlength{\tabcolsep}{2pt}
\renewcommand{\arraystretch}{1.15}
\begin{center}
\resizebox{\textwidth}{!}{%
\begin{tabular}{l|ccccccccccc}
\toprule
\textbf{Method} & T1 & T2 & T3 & T4 & TSExam & Match & Merrill & Anomaly & Cls & TSAQA-DT & Mean \\
\midrule
\rowcolor{bandA}\multicolumn{12}{c}{\textbf{GPT-5.6-luna}} \\
Base~(no tool) & 41.5 & 33.9 & 31.9 & 36.5 & 72.2 & 78.4 & 85.6 & 60.2 & 49.0 & 71.0 & 56.0 \\
\midrule
Chronos-2 fixed & 43.4~{\tiny \textcolor{dpos}{(+1.90)}} & \cellcolor{bestcell}\textbf{49.0}~{\tiny \textcolor{dpos}{(+15.08)}} & 34.8~{\tiny \textcolor{dpos}{(+2.97)}} & \cellcolor{bestcell}\textbf{54.8}~{\tiny \textcolor{dpos}{(+18.25)}} & 76.9~{\tiny \textcolor{dpos}{(+4.70)}} & 77.8~{\tiny \textcolor{dneg}{($-$0.62)}} & 86.4~{\tiny \textcolor{dpos}{(+0.83)}} & 57.2~{\tiny \textcolor{dneg}{($-$3.00)}} & \cellcolor{secondcell}\underline{56.9}~{\tiny \textcolor{dpos}{(+7.92)}} & 69.3~{\tiny \textcolor{dneg}{($-$1.75)}} & \cellcolor{secondcell}\underline{60.7}~{\tiny \textcolor{dpos}{(+4.63)}} \\
TimeART 21 tools & 42.8~{\tiny \textcolor{dpos}{(+1.27)}} & \cellcolor{secondcell}\underline{47.4}~{\tiny \textcolor{dpos}{(+13.49)}} & 26.9~{\tiny \textcolor{dneg}{($-$4.95)}} & 44.4~{\tiny \textcolor{dpos}{(+7.94)}} & \cellcolor{secondcell}\underline{77.6}~{\tiny \textcolor{dpos}{(+5.37)}} & \cellcolor{secondcell}\underline{84.7}~{\tiny \textcolor{dpos}{(+6.25)}} & \cellcolor{bestcell}\textbf{87.2}~{\tiny \textcolor{dpos}{(+1.67)}} & 53.9~{\tiny \textcolor{dneg}{($-$6.25)}} & 44.0~{\tiny \textcolor{dneg}{($-$5.00)}} & 71.8~{\tiny \textcolor{dpos}{(+0.75)}} & 58.1~{\tiny \textcolor{dpos}{(+2.05)}} \\
Self-Refine & 42.2~{\tiny \textcolor{dpos}{(+0.63)}} & 30.0~{\tiny \textcolor{dneg}{($-$3.97)}} & 30.4~{\tiny \textcolor{dneg}{($-$1.49)}} & 27.8~{\tiny \textcolor{dneg}{($-$8.73)}} & 71.6~{\tiny \textcolor{dneg}{($-$0.67)}} & 80.3~{\tiny \textcolor{dpos}{(+1.88)}} & \cellcolor{bestcell}\textbf{87.2}~{\tiny \textcolor{dpos}{(+1.67)}} & \cellcolor{secondcell}\underline{61.4}~{\tiny \textcolor{dpos}{(+1.25)}} & 49.4~{\tiny \textcolor{dpos}{(+0.42)}} & 70.8~{\tiny \textcolor{dneg}{($-$0.25)}} & 55.1~{\tiny \textcolor{dneg}{($-$0.93)}} \\
vote@5 & 39.0~{\tiny \textcolor{dneg}{($-$2.53)}} & 34.7~{\tiny \textcolor{dpos}{(+0.79)}} & 32.9~{\tiny \textcolor{dpos}{(+0.99)}} & 40.5~{\tiny \textcolor{dpos}{(+3.97)}} & 72.9~{\tiny \textcolor{dpos}{(+0.67)}} & 80.3~{\tiny \textcolor{dpos}{(+1.88)}} & 86.8~{\tiny \textcolor{dpos}{(+1.25)}} & 59.7~{\tiny \textcolor{dneg}{($-$0.50)}} & 49.4~{\tiny \textcolor{dpos}{(+0.42)}} & 72.0~{\tiny \textcolor{dpos}{(+1.00)}} & 56.8~{\tiny \textcolor{dpos}{(+0.79)}} \\
few-shot ICL $k{=}4$ & \cellcolor{bestcell}\textbf{48.5}~{\tiny \textcolor{dpos}{(+6.96)}} & 33.9~{\tiny \textcolor{dzero}{(0.00)}} & \cellcolor{bestcell}\textbf{42.8}~{\tiny \textcolor{dpos}{(+10.89)}} & 38.1~{\tiny \textcolor{dpos}{(+1.59)}} & 76.9~{\tiny \textcolor{dpos}{(+4.70)}} & 81.6~{\tiny \textcolor{dpos}{(+3.12)}} & 86.4~{\tiny \textcolor{dpos}{(+0.83)}} & 56.7~{\tiny \textcolor{dneg}{($-$3.50)}} & 39.8~{\tiny \textcolor{dneg}{($-$9.17)}} & \cellcolor{secondcell}\underline{73.0}~{\tiny \textcolor{dpos}{(+2.00)}} & 57.8~{\tiny \textcolor{dpos}{(+1.74)}} \\
\MODEL~(ours) & \cellcolor{secondcell}\underline{44.7}~{\tiny \textcolor{dpos}{(+3.16)}} & 45.0~{\tiny \textcolor{dpos}{(+11.11)}} & \cellcolor{secondcell}\underline{37.2}~{\tiny \textcolor{dpos}{(+5.28)}} & \cellcolor{secondcell}\underline{47.1}~{\tiny \textcolor{dpos}{(+10.58)}} & \cellcolor{bestcell}\textbf{77.8}~{\tiny \textcolor{dpos}{(+5.59)}} & \cellcolor{bestcell}\textbf{90.9}~{\tiny \textcolor{dpos}{(+12.50)}} & \cellcolor{secondcell}\underline{87.1}~{\tiny \textcolor{dpos}{(+1.53)}} & \cellcolor{bestcell}\textbf{74.9}~{\tiny \textcolor{dpos}{(+14.69)}} & \cellcolor{bestcell}\textbf{61.1}~{\tiny \textcolor{dpos}{(+12.08)}} & \cellcolor{bestcell}\textbf{82.4}~{\tiny \textcolor{dpos}{(+11.42)}} & \cellcolor{bestcell}\textbf{64.8}~{\tiny \textcolor{dpos}{(+8.79)}} \\
\midrule
\rowcolor{bandB}\multicolumn{12}{c}{\textbf{GPT-5.6-terra}} \\
Base~(no tool) & 37.9 & 34.0 & 32.7 & 37.2 & 73.6 & 78.3 & 87.4 & 59.8 & 54.0 & 71.5 & 56.6 \\
\midrule
Chronos-2 fixed & 37.9~{\tiny \textcolor{dzero}{(0.00)}} & \cellcolor{bestcell}\textbf{50.7}~{\tiny \textcolor{dpos}{(+16.67)}} & 32.2~{\tiny \textcolor{dneg}{($-$0.50)}} & \cellcolor{bestcell}\textbf{48.3}~{\tiny \textcolor{dpos}{(+11.11)}} & 76.3~{\tiny \textcolor{dpos}{(+2.68)}} & \cellcolor{secondcell}\underline{80.2}~{\tiny \textcolor{dpos}{(+1.88)}} & 88.2~{\tiny \textcolor{dpos}{(+0.83)}} & 61.8~{\tiny \textcolor{dpos}{(+2.00)}} & 52.8~{\tiny \textcolor{dneg}{($-$1.27)}} & \cellcolor{secondcell}\underline{73.8}~{\tiny \textcolor{dpos}{(+2.25)}} & \cellcolor{secondcell}\underline{60.2}~{\tiny \textcolor{dpos}{(+3.56)}} \\
TimeART 21 tools & 41.0~{\tiny \textcolor{dpos}{(+3.16)}} & 43.5~{\tiny \textcolor{dpos}{(+9.52)}} & 28.2~{\tiny \textcolor{dneg}{($-$4.46)}} & 43.5~{\tiny \textcolor{dpos}{(+6.35)}} & \cellcolor{bestcell}\textbf{78.3}~{\tiny \textcolor{dpos}{(+4.70)}} & 75.2~{\tiny \textcolor{dneg}{($-$3.12)}} & 85.3~{\tiny \textcolor{dneg}{($-$2.08)}} & 51.0~{\tiny \textcolor{dneg}{($-$8.75)}} & 53.6~{\tiny \textcolor{dneg}{($-$0.42)}} & 69.8~{\tiny \textcolor{dneg}{($-$1.75)}} & 56.9~{\tiny \textcolor{dpos}{(+0.31)}} \\
Self-Refine & 38.5~{\tiny \textcolor{dpos}{(+0.63)}} & 31.6~{\tiny \textcolor{dneg}{($-$2.38)}} & \cellcolor{secondcell}\underline{32.7}~{\tiny \textcolor{dzero}{(0.00)}} & 33.2~{\tiny \textcolor{dneg}{($-$3.97)}} & 74.3~{\tiny \textcolor{dpos}{(+0.67)}} & 79.0~{\tiny \textcolor{dpos}{(+0.62)}} & 87.8~{\tiny \textcolor{dpos}{(+0.42)}} & 59.5~{\tiny \textcolor{dneg}{($-$0.25)}} & \cellcolor{secondcell}\underline{54.5}~{\tiny \textcolor{dpos}{(+0.42)}} & 69.5~{\tiny \textcolor{dneg}{($-$2.00)}} & 56.1~{\tiny \textcolor{dneg}{($-$0.58)}} \\
vote@5 & 36.0~{\tiny \textcolor{dneg}{($-$1.90)}} & 32.4~{\tiny \textcolor{dneg}{($-$1.59)}} & \cellcolor{secondcell}\underline{32.7}~{\tiny \textcolor{dzero}{(0.00)}} & 40.3~{\tiny \textcolor{dpos}{(+3.17)}} & 68.9~{\tiny \textcolor{dneg}{($-$4.70)}} & 79.6~{\tiny \textcolor{dpos}{(+1.25)}} & 86.9~{\tiny \textcolor{dneg}{($-$0.42)}} & \cellcolor{secondcell}\underline{64.8}~{\tiny \textcolor{dpos}{(+5.00)}} & 49.9~{\tiny \textcolor{dneg}{($-$4.17)}} & 72.8~{\tiny \textcolor{dpos}{(+1.25)}} & 56.4~{\tiny \textcolor{dneg}{($-$0.21)}} \\
few-shot ICL $k{=}4$ & \cellcolor{bestcell}\textbf{46.1}~{\tiny \textcolor{dpos}{(+8.23)}} & 33.2~{\tiny \textcolor{dneg}{($-$0.79)}} & 29.7~{\tiny \textcolor{dneg}{($-$2.97)}} & 38.8~{\tiny \textcolor{dpos}{(+1.59)}} & \cellcolor{secondcell}\underline{77.0}~{\tiny \textcolor{dpos}{(+3.36)}} & 77.7~{\tiny \textcolor{dneg}{($-$0.62)}} & \cellcolor{secondcell}\underline{89.4}~{\tiny \textcolor{dpos}{(+2.08)}} & 64.3~{\tiny \textcolor{dpos}{(+4.50)}} & 50.7~{\tiny \textcolor{dneg}{($-$3.33)}} & 67.0~{\tiny \textcolor{dneg}{($-$4.50)}} & 57.4~{\tiny \textcolor{dpos}{(+0.76)}} \\
\MODEL~(ours) & \cellcolor{secondcell}\underline{41.7}~{\tiny \textcolor{dpos}{(+3.80)}} & \cellcolor{secondcell}\underline{44.3}~{\tiny \textcolor{dpos}{(+10.32)}} & \cellcolor{bestcell}\textbf{36.1}~{\tiny \textcolor{dpos}{(+3.47)}} & \cellcolor{secondcell}\underline{45.1}~{\tiny \textcolor{dpos}{(+7.94)}} & \cellcolor{secondcell}\underline{77.0}~{\tiny \textcolor{dpos}{(+3.36)}} & \cellcolor{bestcell}\textbf{90.8}~{\tiny \textcolor{dpos}{(+12.50)}} & \cellcolor{bestcell}\textbf{90.3}~{\tiny \textcolor{dpos}{(+2.92)}} & \cellcolor{bestcell}\textbf{72.3}~{\tiny \textcolor{dpos}{(+12.50)}} & \cellcolor{bestcell}\textbf{54.8}~{\tiny \textcolor{dpos}{(+0.83)}} & \cellcolor{bestcell}\textbf{86.0}~{\tiny \textcolor{dpos}{(+14.50)}} & \cellcolor{bestcell}\textbf{63.9}~{\tiny \textcolor{dpos}{(+7.21)}} \\
\midrule
\rowcolor{bandC}\multicolumn{12}{c}{\textbf{GPT-5.4-mini}} \\
Base~(no tool) & 40.9 & 34.3 & 32.5 & 37.6 & 70.8 & 72.1 & 86.1 & 61.0 & 43.1 & 67.3 & 54.6 \\
\midrule
Chronos-2 fixed & 45.4~{\tiny \textcolor{dpos}{(+4.43)}} & \cellcolor{bestcell}\textbf{47.8}~{\tiny \textcolor{dpos}{(+13.49)}} & 28.5~{\tiny \textcolor{dneg}{($-$3.96)}} & \cellcolor{bestcell}\textbf{59.8}~{\tiny \textcolor{dpos}{(+22.22)}} & 70.2~{\tiny \textcolor{dneg}{($-$0.67)}} & 69.0~{\tiny \textcolor{dneg}{($-$3.12)}} & 84.0~{\tiny \textcolor{dneg}{($-$2.08)}} & 56.2~{\tiny \textcolor{dneg}{($-$4.75)}} & 45.2~{\tiny \textcolor{dpos}{(+2.08)}} & 66.8~{\tiny \textcolor{dneg}{($-$0.50)}} & \cellcolor{secondcell}\underline{57.3}~{\tiny \textcolor{dpos}{(+2.71)}} \\
TimeART 21 tools & 38.4~{\tiny \textcolor{dneg}{($-$2.53)}} & \cellcolor{secondcell}\underline{46.2}~{\tiny \textcolor{dpos}{(+11.90)}} & 31.0~{\tiny \textcolor{dneg}{($-$1.49)}} & \cellcolor{secondcell}\underline{51.9}~{\tiny \textcolor{dpos}{(+14.29)}} & \cellcolor{bestcell}\textbf{76.9}~{\tiny \textcolor{dpos}{(+6.04)}} & 71.5~{\tiny \textcolor{dneg}{($-$0.62)}} & 84.9~{\tiny \textcolor{dneg}{($-$1.25)}} & 55.2~{\tiny \textcolor{dneg}{($-$5.79)}} & \cellcolor{secondcell}\underline{48.5}~{\tiny \textcolor{dpos}{(+5.42)}} & 65.1~{\tiny \textcolor{dneg}{($-$2.25)}} & 56.9~{\tiny \textcolor{dpos}{(+2.37)}} \\
Self-Refine & 41.6~{\tiny \textcolor{dpos}{(+0.63)}} & 32.7~{\tiny \textcolor{dneg}{($-$1.59)}} & 32.0~{\tiny \textcolor{dneg}{($-$0.50)}} & 37.6~{\tiny \textcolor{dzero}{(0.00)}} & 66.8~{\tiny \textcolor{dneg}{($-$4.03)}} & \cellcolor{secondcell}\underline{80.2}~{\tiny \textcolor{dpos}{(+8.12)}} & 86.5~{\tiny \textcolor{dpos}{(+0.42)}} & 60.2~{\tiny \textcolor{dneg}{($-$0.75)}} & 46.5~{\tiny \textcolor{dpos}{(+3.38)}} & \cellcolor{secondcell}\underline{68.3}~{\tiny \textcolor{dpos}{(+1.00)}} & 55.2~{\tiny \textcolor{dpos}{(+0.67)}} \\
vote@5 & 41.6~{\tiny \textcolor{dpos}{(+0.63)}} & 34.3~{\tiny \textcolor{dzero}{(0.00)}} & \cellcolor{secondcell}\underline{34.5}~{\tiny \textcolor{dpos}{(+1.98)}} & 39.2~{\tiny \textcolor{dpos}{(+1.59)}} & 69.5~{\tiny \textcolor{dneg}{($-$1.34)}} & 75.8~{\tiny \textcolor{dpos}{(+3.75)}} & \cellcolor{secondcell}\underline{88.6}~{\tiny \textcolor{dpos}{(+2.50)}} & \cellcolor{secondcell}\underline{63.0}~{\tiny \textcolor{dpos}{(+2.00)}} & 47.7~{\tiny \textcolor{dpos}{(+4.58)}} & 68.1~{\tiny \textcolor{dpos}{(+0.75)}} & 56.2~{\tiny \textcolor{dpos}{(+1.64)}} \\
few-shot ICL $k{=}4$ & \cellcolor{bestcell}\textbf{49.2}~{\tiny \textcolor{dpos}{(+8.23)}} & 29.5~{\tiny \textcolor{dneg}{($-$4.76)}} & 30.5~{\tiny \textcolor{dneg}{($-$1.98)}} & 28.8~{\tiny \textcolor{dneg}{($-$8.73)}} & \cellcolor{secondcell}\underline{72.9}~{\tiny \textcolor{dpos}{(+2.01)}} & 70.8~{\tiny \textcolor{dneg}{($-$1.25)}} & \cellcolor{bestcell}\textbf{89.0}~{\tiny \textcolor{dpos}{(+2.92)}} & \cellcolor{secondcell}\underline{63.0}~{\tiny \textcolor{dpos}{(+2.00)}} & 32.7~{\tiny \textcolor{dneg}{($-$10.42)}} & 67.8~{\tiny \textcolor{dpos}{(+0.50)}} & 53.4~{\tiny \textcolor{dneg}{($-$1.15)}} \\
\MODEL~(ours) & \cellcolor{secondcell}\underline{46.6}~{\tiny \textcolor{dpos}{(+5.70)}} & 45.4~{\tiny \textcolor{dpos}{(+11.11)}} & \cellcolor{bestcell}\textbf{37.0}~{\tiny \textcolor{dpos}{(+4.46)}} & 39.2~{\tiny \textcolor{dpos}{(+1.59)}} & \cellcolor{secondcell}\underline{72.9}~{\tiny \textcolor{dpos}{(+2.01)}} & \cellcolor{bestcell}\textbf{89.0}~{\tiny \textcolor{dpos}{(+16.88)}} & \cellcolor{bestcell}\textbf{89.0}~{\tiny \textcolor{dpos}{(+2.92)}} & \cellcolor{bestcell}\textbf{77.2}~{\tiny \textcolor{dpos}{(+16.25)}} & \cellcolor{bestcell}\textbf{53.5}~{\tiny \textcolor{dpos}{(+10.42)}} & \cellcolor{bestcell}\textbf{83.3}~{\tiny \textcolor{dpos}{(+16.00)}} & \cellcolor{bestcell}\textbf{63.3}~{\tiny \textcolor{dpos}{(+8.73)}} \\
\midrule
\rowcolor{bandD}\multicolumn{12}{c}{\textbf{Strong Models + the Library Evolved by GPT-5.6-luna}} \\
Claude-opus-5~(bare) & 36.1 & 38.9 & 30.2 & 41.3 & 74.5 & 72.5 & 91.7 & 60.2 & 51.2 & 59.8 & 55.6 \\
\rowcolor{bestcell}\quad + evolved library & 40.5~{\tiny \textcolor{dpos}{(+4.43)}} & 40.5~{\tiny \textcolor{dpos}{(+1.59)}} & 34.2~{\tiny \textcolor{dpos}{(+3.96)}} & 46.8~{\tiny \textcolor{dpos}{(+5.56)}} & 81.2~{\tiny \textcolor{dpos}{(+6.71)}} & 96.2~{\tiny \textcolor{dpos}{(+23.75)}} & 94.2~{\tiny \textcolor{dpos}{(+2.50)}} & 86.5~{\tiny \textcolor{dpos}{(+26.25)}} & 65.0~{\tiny \textcolor{dpos}{(+13.75)}} & 76.2~{\tiny \textcolor{dpos}{(+16.50)}} & 66.1~{\tiny \textcolor{dpos}{(+10.50)}} \\
Claude-sonnet-5~(bare) & 38.0 & 38.1 & 30.2 & 42.9 & 71.1 & 76.9 & 91.7 & 59.0 & 50.8 & 60.0 & 55.9 \\
\rowcolor{bestcell}\quad + evolved library & 41.8~{\tiny \textcolor{dpos}{(+3.80)}} & 43.6~{\tiny \textcolor{dpos}{(+5.56)}} & 33.2~{\tiny \textcolor{dpos}{(+2.97)}} & 44.4~{\tiny \textcolor{dpos}{(+1.59)}} & 83.2~{\tiny \textcolor{dpos}{(+12.08)}} & 96.2~{\tiny \textcolor{dpos}{(+19.38)}} & 94.2~{\tiny \textcolor{dpos}{(+2.50)}} & 85.5~{\tiny \textcolor{dpos}{(+26.50)}} & 66.2~{\tiny \textcolor{dpos}{(+15.42)}} & 76.8~{\tiny \textcolor{dpos}{(+16.75)}} & 66.5~{\tiny \textcolor{dpos}{(+10.65)}} \\
GPT-5.6-sol~(bare) & 38.0 & 30.9 & 30.7 & 35.7 & 75.8 & 80.6 & 90.0 & 59.5 & 57.9 & 71.8 & 57.1 \\
\rowcolor{bestcell}\quad + evolved library & 43.7~{\tiny \textcolor{dpos}{(+5.70)}} & 39.7~{\tiny \textcolor{dpos}{(+8.73)}} & 36.1~{\tiny \textcolor{dpos}{(+5.45)}} & 43.6~{\tiny \textcolor{dpos}{(+7.94)}} & 81.9~{\tiny \textcolor{dpos}{(+6.04)}} & 95.0~{\tiny \textcolor{dpos}{(+14.37)}} & 92.9~{\tiny \textcolor{dpos}{(+2.92)}} & 81.2~{\tiny \textcolor{dpos}{(+21.75)}} & 55.4~{\tiny \textcolor{dneg}{($-$2.50)}} & 84.0~{\tiny \textcolor{dpos}{(+12.25)}} & 65.4~{\tiny \textcolor{dpos}{(+8.27)}} \\
\bottomrule
\end{tabular}}
\end{center}
\end{table}

\subsection{Experimental Settings}
\label{sec:setup}

\textbf{Tasks.}
The formal suite covers ten time series QA tasks from six public sources: TemporalBench T1--T4~\citep{weng2026temporalbench}, TimeSeriesExam~\citep{cai2024timeseriesexam}, Merrill~\citep{merrill2024language}, TimeMQA Anomaly and Classification~\citep{kong2025time}, MMTS Match~\citep{yin2026mmts}, and TSAQA Data-Transformation~\citep{jing2026tsaqa}.
Split details are in Appendix~\ref{app:dataset_details}.

\textbf{Protocol.}
Every run uses the same frozen protocol: empty root library, one evolution round, train-only calibration, and a report-only test split.
We evaluate on three backbones, GPT-5.6-luna, GPT-5.6-terra, and GPT-5.4-mini, with all four LM roles switched together.
Every number we report is a within-run paired delta~(\cref{eq:paired}).

\textbf{Baselines.}
We compare against five non-evolving baselines: a fixed Chronos-2 forecasting tool~\citep{ansari2025chronos}, the 21 expert-curated tools of TimeART~\citep{wu2026timeart}, Self-Refine~\citep{madaan2023self}, five-sample self-consistency voting, and few-shot in-context learning with four training examples.
Each tests one alternative explanation of the gains: that a strong pre-installed tool, a hand-engineered library, one more round of reflection, the same budget spent on sampling, or direct contact with the training data would suffice.
All are measured with the same paired protocol on the same splits.

\subsection{Overall Comparison}
\label{sec:main}

\begin{figure}[t]
\centering
\includegraphics[width=\textwidth]{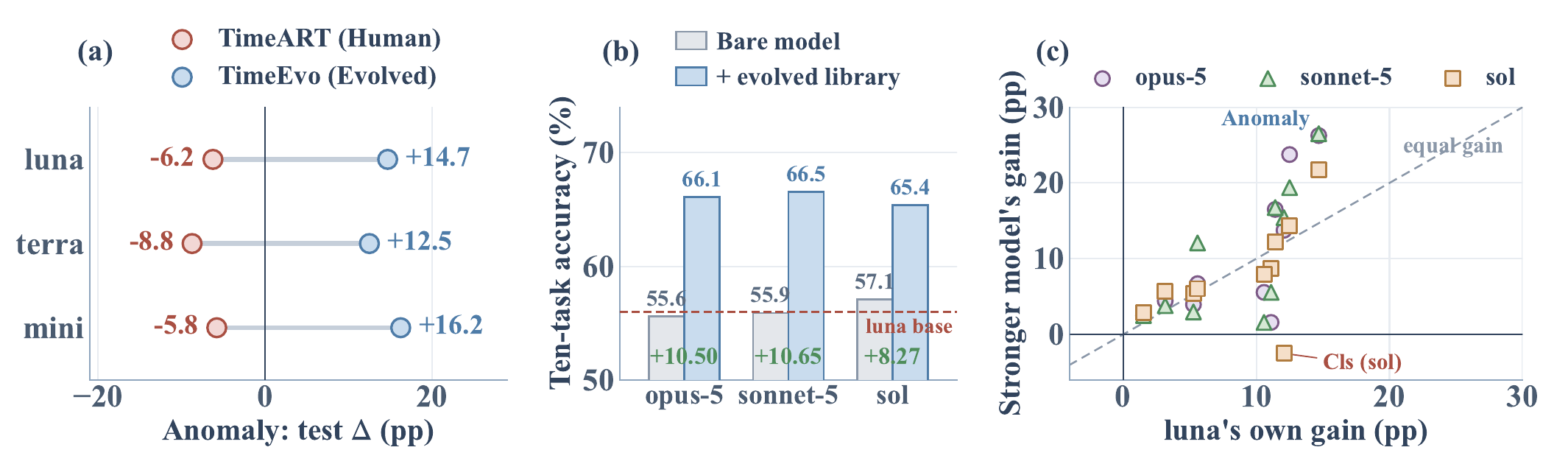}
\caption{
Human tools, and the transfer of an evolved library.
\textbf{(a)}:~Test delta on TimeMQA Anomaly for the 21 expert-curated tools of TimeART and for the evolved library.
\textbf{(b)}:~Ten-task mean accuracy of three stronger models, carrying the library GPT-5.6-luna evolved.
\textbf{(c)}:~Per-task gain of those models against the gain luna itself obtained with the same library. Diagonal marks equal gain.
}
\label{fig:whybaselines}
\vspace{-8pt}
\end{figure}

We compare \MODEL against the five baselines on all ten tasks, under each of the three backbones.
Results are in~\cref{tab:main}, and~\cref{fig:whybaselines}a takes a closer look at TimeMQA Anomaly, where the human library fails.
Based on these results, we summarize our observations~(\textbf{Obs.}) as follows:

\textbf{Obs.~\ding{182}: \MODEL is the only method that improves every task on every backbone.}
All thirty of its task results are positive.
Its mean gain is the largest in every block: $+8.79$ on luna, $+8.73$ on mini, and $+7.21$ on terra, the strongest backbone of the three.
Each backbone evolves its own library from its own failures, so the gains come from the loop rather than from one lucky set of tools.
These results answer the first half of the central question of~\cref{sec:intro}: an agent can supply its own missing tools from its own diagnosed failures.

\textbf{Obs.~\ding{183}: Human-supplied tools and generic self-improvement pay for their peaks with losses elsewhere.}
The fixed Chronos-2 tool is strongly positive only on the two forecasting tiers it was built for~(T2/T4, up to $+22.22$) and flat or negative elsewhere.
As shown in~\cref{fig:whybaselines}a, the TimeART library includes a dedicated anomaly tool yet loses $-6.25/-8.75/-5.79$ on Anomaly under every backbone, and one Self-Refine pass breaks nearly as many answers as it fixes~(\cref{fig:intro}b).
Voting with five samples spends five times the budget for ten-task means of $+0.79/-0.21/+1.64$, and few-shot ICL, which reads the same training split our method learns from, is strongly negative on Classification under all three backbones.
None of these alternatives is free of human design effort.

\subsection{Transfer Analysis}
\label{sec:transfer}

We test whether a library evolved on one model still helps a different one.
For each task we install the library that GPT-5.6-luna evolved into Claude-opus-5, Claude-sonnet-5, and GPT-5.6-sol, and no test data is used to pick it.
From the last block of~\cref{tab:main} and from~\cref{fig:whybaselines}b-c, we observe:

\textbf{Obs.~\ding{184}: A stronger model is not a substitute for an evolved library.}
As shown in~\cref{fig:whybaselines}b, without tools the three expensive models score no better than the cheapest one, with everything between 54.6 and 57.1.
Installing luna's library lifts each of them on the ten-task mean by $+10.50$, $+10.65$, and $+8.27$, and only one of the thirty task results is negative.
As a result, we can conclude that what they were missing was not a better model but the right tools, and that what the library carries is task knowledge rather than model-specific habits.

\textbf{Obs.~\ding{185}: The library gains more on models stronger than the one that grew it.}
As shown in~\cref{fig:whybaselines}c, 21 of the 30 task results sit above the diagonal, so the same library usually gains more on the stronger model than it did on luna.
On Anomaly the two Claude models gain $+26.25$ and $+26.50$, against $+14.69$ on luna itself.
The tools compute the same evidence in every case, and what differs is how deeply a model can exploit it.
Therefore, growing the tools need not happen on an expensive model at all.

\begin{figure}[t]
\centering
\includegraphics[width=\textwidth]{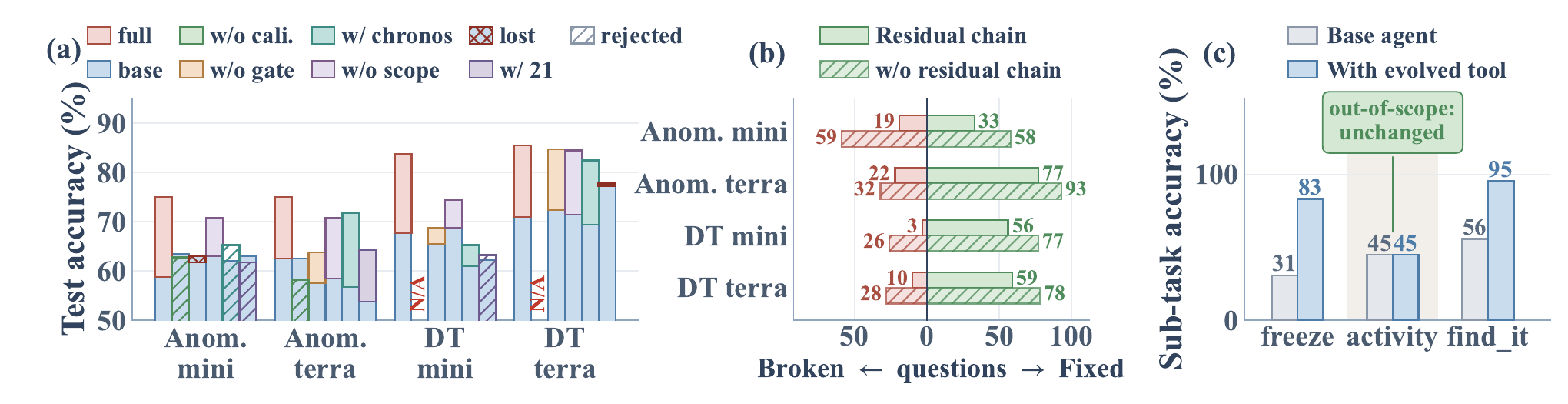}
\caption{
Ablations on TimeMQA Anomaly and TSAQA-DT under two backbones.
\textbf{(a)}:~Test accuracy with one component removed, each bar rising from that run's own no-tool base~(blue).
A red cross marks accuracy the library lost, and a hatched bar marks a gate-rejected round.
\textbf{(b)}:~Answers fixed and broken by the update, under the protected review and under unconditional overwrite of the same library.
\textbf{(c)}:~Sub-task accuracy with and without the evolved tool, on two in-scope sub-tasks and one out-of-scope~(shaded).
Full numbers in Appendix~\ref{sec:supp}.
}
\label{fig:ablation}
\end{figure}

\subsection{Ablation and General Analysis}
\label{sec:ablation}

\begin{figure}[t]
\centering
\includegraphics[width=\textwidth]{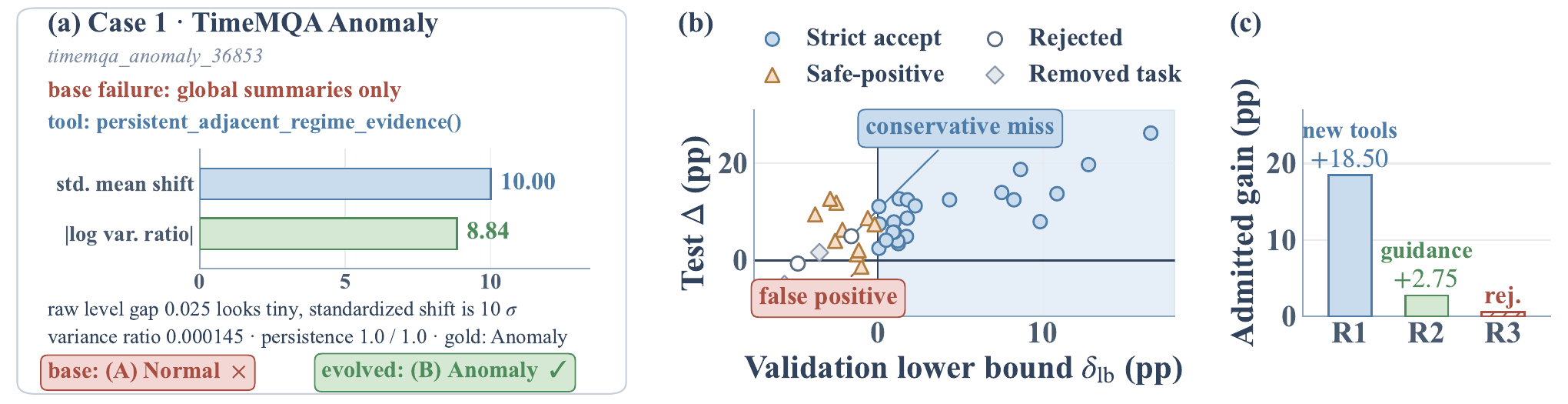}
\caption{
What the agent builds, and how the gate behaves.
\textbf{(a)}:~One question from the report-only test trace: the tool the agent built, the numbers it returned, and the answer before and after.
\textbf{(b)}:~Gate calibration across all formal runs, validation lower bound against deployed test delta, with the report-only diagnostic for rejected rounds.
\textbf{(c)}:~Admitted gain per round on a three-round run~(terra, Anomaly).
Two further cases in Appendix~\ref{app:cases}.
}
\label{fig:cards}
\end{figure}

We first remove one component at a time, then open the deployed libraries and check how the gate behaves.
We ablate on two datasets, Anomaly and TSAQA-DT, each under mini and terra, and the four cells were chosen before the runs.
Against the full method we compare six changes, named as in~\cref{fig:ablation}.
(\textbf{1})~\textbf{w/o cali.} drops the train-only decision trees, which only the Anomaly libraries carry.
(\textbf{2})~\textbf{w/o gate} force-accepts the candidate library.
(\textbf{3})~\textbf{w/o scope} removes the structured scopes.
(\textbf{4})~\textbf{w/ chronos} and (\textbf{5})~\textbf{w/ 21 tools} replace the empty root by a pre-installed forecasting tool and by the 21 expert-curated tools of TimeART.
(\textbf{6})~\textbf{w/o residual chain} overwrites the base answers unconditionally instead of reviewing them.
From~\cref{fig:ablation,fig:cards}, we observe:

\textbf{Obs.~\ding{186}: Every component earns its place, and none can be removed for free.}
The full method reaches the highest test accuracy in all four cells.
Calibration and the gate cost the most: without calibration both Anomaly runs deploy nothing, and without the gate the mini run deploys a library that ends $1.25$ below its own base, where the full method ends $16.25$ above.
Removing structured scopes costs $1.00$ to $9.25$ points.
In~\cref{fig:ablation}b, overwriting fixes 16 to 25 more answers per cell than reviewing does, and breaks far more: 59 versus 19, 32 versus 22, 26 versus 3, and 28 versus 10.
The chain adds no repairs itself, but keeps a good library from breaking answers that were already right.

\textbf{Obs.~\ding{187}: A pre-installed root is not a shortcut.}
As shown in~\cref{fig:ablation}a, both alternative roots end below the empty root in all four cells: the Chronos wrapper by $3.04$ to $18.50$ points and the 21 expert-curated tools of TimeART by $8.25$ to $21.50$.
The human library is the weaker start of the two, and evolution from it is rejected by the gate on both mini tasks, and on TSAQA-DT under terra it ends $0.50$ below its own base.
Tools written before deployment do not become the right tools when an agent evolves on top of them, which is the \textbf{Human--Agent Tool Misalignment} of Phenomenon~\ding{182} appearing inside our own method.

\textbf{Obs.~\ding{188}: The admitted tools are targeted and used exactly where they claim.}
As illustrated in~\cref{fig:ablation}c, on TimeMQA Classification the deployed tool computes exactly the band powers that the question's criteria name, and the out-of-scope activity sub-task stays at exactly 45.0.
On TSAQA-DT, the libraries evolved under all three backbones independently rediscover the same Fourier and wavelet family, so the measurement contract, not a lucky sample, determines what gets built. 25 of the 29 deployed libraries are pure tools, and the other four admitted a textual patch, because the gate arbitrates by measured repair and harm rather than by artifact type.

\textbf{Obs.~\ding{189}: The repairs are single missing measurements, not better reasoning.}
As shown in~\cref{fig:cards}a, the base agent fails this question without computing anything, reading a global summary of a series whose overall spread looks ordinary.
The evolved tool supplies exactly one measurement and the answer flips.
The raw level gap between the two adjacent windows is $0.025$, but standardized it is $10\,\sigma$, and the variance ratio of $0.000145$ says the second window is almost flat.
The two cases in Appendix~\ref{app:cases} fail the same way: the base agent is not bad at arithmetic but simply never does it.

\textbf{Obs.~\ding{190}: The gate errs visibly, in both directions, and the loop stops itself.}
As shown in~\cref{fig:cards}b, every strict-region acceptance~($\delta_{\mathrm{lb}} > 0$) lands test-positive. 
The borderline band contains one false positive~(a T1 library at $-1.27$, the only accepted library that lands test-negative), and the rejected region contains one conservative false negative~(a Merrill library whose report-only diagnostic scored $+5.00$).
A rejected round rolls back cleanly and counts as zero.
As illustrated in~\cref{fig:cards}c, on a three-round run, round 1 admits new tools~(validation gain $+18.50$), round 2 admits only usage guidance~($+2.75$), and round 3 is rejected outright. Every tool in the final library carries a round-1 identifier, so the tools stabilize in one round and later rounds only teach how to use them.

\section{Conclusion}
\label{sec:conclusion}

This paper studies two failures of tool-based time series agents: the tools humans supply are not the ones the agent needs at runtime, and an agent's own revisions break answers that the final score hides.
To address them, we introduce \MODEL, which turns diagnosed failures into evidence-only tools and admits a candidate library only when a paired comparison shows it fixes more than it breaks, rolling the round back otherwise.
Experiments on ten time series QA tasks and three backbones show that \MODEL, starting from an empty library, improves accuracy on every task and every backbone, and that a library grown on a cheap model still gains when it is installed into stronger ones.
An agent's own failures are therefore a better specification for tools than human anticipation.

\bibliography{iclr2027_conference}
\bibliographystyle{iclr2027_conference}

\clearpage

\appendix
\section{Dataset Details}
\label{app:dataset_details}

\begin{table}[h]
\footnotesize
\caption{The ten-task suite: data sources and frozen train/validation/test sizes. Merrill is downsampled from its official 5779/1000/1036 split, and every other task uses its full frozen pool. Splits are grouped by source series, so questions from the same series never cross splits.}
\label{tab:splits}
\begin{center}
\begin{tabular*}{\textwidth}{@{\extracolsep{\fill}}llrrr@{}}
\toprule
Task & Source & Train & Val & Test \\
\midrule
T1 & TemporalBench~\citep{weng2026temporalbench} & 400 & 156 & 158 \\
T2 & TemporalBench~\citep{weng2026temporalbench} & 324 & 123 & 126 \\
T3 & TemporalBench~\citep{weng2026temporalbench} & 517 & 196 & 202 \\
T4 & TemporalBench~\citep{weng2026temporalbench} & 324 & 123 & 126 \\
TSExam & TimeSeriesExam~\citep{cai2024timeseriesexam} & 448 & 149 & 149 \\
Merrill & Time-Series Reasoning~\citep{merrill2024language} & 240 & 120 & 240 \\
Anomaly & Time-MQA~\citep{kong2025time} & 400 & 400 & 400 \\
Classification & Time-MQA~\citep{kong2025time} & 240 & 120 & 240 \\
Match & MMTS-Bench~\citep{yin2026mmts} & 160 & 80 & 160 \\
TSAQA-DT & TSAQA~\citep{jing2026tsaqa} & 400 & 400 & 400 \\
\bottomrule
\end{tabular*}
\end{center}
\end{table}

The ten tasks come from six public sources, and every source is adapted to one row contract: a question, a closed option set whose gold answer appears verbatim among the options, one or two primary series, and optional named covariate or candidate series.
This appendix describes each source and the split design, and \cref{tab:splits} lists the frozen sizes.

\textbf{TemporalBench~(T1--T4)~\citep{weng2026temporalbench}.}
The four tiers share one source pool and are split by source sample, so all questions expanded from the same sample stay in the same split.
T1 asks property questions about the observed history~(trend, volatility, seasonality, anomaly), with label spaces read from the source prompts rather than invented.
T2 and T4 are forecasting-style multiple-choice questions, distinguished only by whether event context is present, and T3 packs several sub-questions per sample with clinical covariates such as temperature and respiratory rate.
All series are univariate with a median length of 300 points, and only the observed history is ever exposed, never another tier's future targets.

\textbf{TimeSeriesExam~\citep{cai2024timeseriesexam}.}
An exam of time series knowledge and the only task with two-series questions~(99 single and 50 dual in the test split).
Its series are the longest in the suite~(median 1024 points).
The split groups all questions that share the exact same input series and stratifies by question type, difficulty, option count, and even the position of the correct option, using no model outputs.

\textbf{Merrill~\citep{merrill2024language}.}
Scenario matching: every question shows one series and four candidate natural-language scenario descriptions, under one shared question text.
We keep the official train/validation/test files and downsample them to 240/120/240 for cost. Series lengths span 12 to 1460 points, the widest range in the suite.

\textbf{Time-MQA~(Anomaly and Classification)~\citep{kong2025time}.}
The source embeds each series inside the question text. We extract the longest numeric list into a machine-readable series so tools can compute on it, and turn the closed-set free-text answer into explicit options.
From 37{,}000 source rows, one representative row is kept per unique series, and one group whose duplicate series carried contradictory labels is excluded entirely.
Anomaly is a two-way normal-versus-anomaly task with the shortest series in the suite~(8 to 64 points, median 16).
Classification mixes a six-way activity-recognition half with a two-way freeze-of-gait half, over accelerometer snippets of 9 to 30 points.

\textbf{MMTS Match~\citep{yin2026mmts}.}
Each question shows a target series and four candidate series and asks which candidate is most similar.
The four candidates are carried as named covariate series, and the task is kept as a four-way choice rather than rewritten into easier pairwise comparisons.
The 400 source rows contain only 112 unique target series, so whole target groups are assigned to one split.

\textbf{TSAQA~(Data-Transformation)~\citep{jing2026tsaqa}.}
Closed-set questions about transformed series~(multiple-choice and true-or-false halves), with options lifted from the source text and one series per question.
Row identifiers encode the official split, so our frozen manifest never crosses the official train/validation/test boundaries.

\textbf{Leakage control and heterogeneity.}
Across the suite, splits are grouped at the series level~(all questions sharing a series move together), duplicate series are collapsed to one representative, contradictory groups are dropped, official split boundaries are preserved, and future windows are never exposed to the agent or its tools.
Series lengths span two orders of magnitude~(8 to 1460 points) and option counts range from two to six, so absolute accuracies are not comparable across tasks. All comparisons in this paper are therefore within-task paired deltas.

\section{Additional Case Studies}
\label{app:cases}

\begin{figure}[h]
\centering
\includegraphics[width=\textwidth]{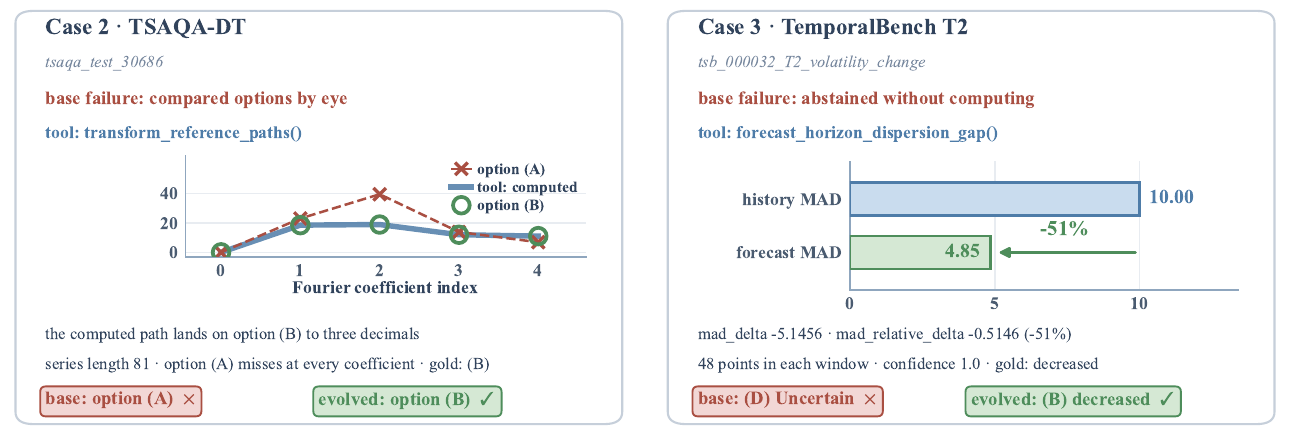}
\caption{
Two further repairs.
\textbf{Left}: the agent picked the Fourier spectrum that looked right, and the synthesized tool computed the real one, which matches the correct option to three decimals.
\textbf{Right}: the agent abstained because it never compared the two windows, and the tool measured the dispersion drop directly.
}
\label{fig:cards_appendix}
\end{figure}

Both cases below are single questions from the report-only test traces of accepted rounds, quoted with the tool's own returned values, and both are drawn in~\cref{fig:cards_appendix}.
For each case we give the question, the answer before and after the round, the failure the planner diagnosed, the measurement contract it wrote, and the numbers the synthesized tool returned on that question.

\textbf{Comparing options by eye.}
On TSAQA-DT~(\texttt{tsaqa\_test\_30686}, terra), the question asks which of the listed option vectors is the Fourier transform of the given series.
The base agent answers~(A), whose leading coefficients $[0, 23.09, 39.60, 13.86, 6.90]$ are the largest and therefore look the most like a spectrum, and the gold answer is~(B), $[0, 18.55, 18.96, 12.00, 11.07]$.
The planner diagnoses the bucket as a candidate-comparison error, namely that the solver never produces aligned quantitative comparisons between the requested transform and the candidates, and writes a contract for candidate-wise normalized residuals after computing the requested representation and aligning each candidate by valid length and coefficient index.
The synthesized tool \texttt{transform\_reference\_paths} returns a series length of 81 and a magnitude path of $[0.0005, 18.5500, 18.9578, 12.0046, 11.0670, 7.5275, \ldots]$, which matches option~(B) to three decimals at every index, and the evolved agent answers~(B).

\textbf{Abstaining without computing.}
On TemporalBench T2~(\texttt{tsb\_000032\_T2\_volatility\_change}, mini), the question gives a history window and asks whether volatility over the forecast horizon rises, falls, or cannot be determined.
The base agent answers~(D) Uncertain, which is the fallback option rather than a wrong measurement, and the gold answer is~(B) decreased.
The planner diagnoses that the agent never computed the cross-window spread comparison the question needs, and writes a contract for a deterministic comparison of forecast-horizon dispersion against historical dispersion.
The synthesized tool \texttt{forecast\_horizon\_dispersion\_gap} returns a history MAD of $10.0000$ against a forecast MAD of $4.8544$, a delta of $-5.1456$, a relative delta of $-0.5146$, and 48 points in each window, and the evolved agent answers~(B).

Neither repair required better reasoning about time series.
In both cases the agent had already read the question correctly and simply had no number to read off, and one deterministic measurement was enough to settle the answer.

\section{Frozen Protocol and Round Procedure}
\label{sec:config}

This appendix records the complete frozen configuration and the step-by-step round procedure for reproducibility.

\subsection{Round Procedure}
\label{sec:procedure}

\begin{enumerate}\setcounter{enumi}{-1}
  \item \textbf{Root library.} The root is configurable and immutable. \texttt{empty} installs no tool and is the frozen protocol used for every reported result, while \texttt{chronos} installs a Chronos-2 wrapper and is retained only as an ablation.

  \item \textbf{Base pass.} The current chain answers the full training split under a semantic-fingerprint cache. This yields the base accuracy of the round and the error set that everything downstream is built from.

  \item \textbf{Failure clustering.} Errors are partitioned by whether a tool fired on them, and each partition is clustered independently by a triage LM in two stages. A clustering failure invalidates the round as measurement-invalid, which is a rollback rather than a quality verdict.
\begin{itemize}[leftmargin=1.2em, topsep=2pt, itemsep=1pt, parsep=0pt]
    \item \textbf{Partition.} The \texttt{tool\_fired} side collects questions where a tool ran and the answer was still wrong, which points at a defective or misused tool, and the \texttt{no\_tool} side collects questions no tool reached, which points at a coverage gap. An empty root puts every error on the second side.
    \item \textbf{Define.} The stage samples at most 150 errors and returns at most eight categories, each carrying a short label and a criteria string, with at most two retries.
    \item \textbf{Assign.} Every error is sent in batches of 32 under a strict JSON contract that must cover each id in the batch, with at most two retries per batch and at most three passes over what remains. Small batches are deliberate, since one long response that fails validation would otherwise void a large batch.
    \item \textbf{Resplit and drop.} A bucket larger than 64 errors is split once and not recursively. Errors that stay unassigned, and errors in buckets with fewer than two members, are dropped.
\end{itemize}

  \item \textbf{Two-level planning.} A local planner diagnoses each bucket independently, and a global planner then coordinates the proposals before anything is synthesized.
\begin{itemize}[leftmargin=1.2em, topsep=2pt, itemsep=1pt, parsep=0pt]
    \item \textbf{Local planner.} It sees the bucket's error dossier, at most 12 support examples, and at most 48 background examples, and returns one proposal card.
    \item \textbf{Descriptive fields.} The action~(\textsc{\textbf{create}}, \textsc{\textbf{refine}}, \textsc{\textbf{rescope}}, or \textsc{\textbf{retire}}), the artifact kind, the target tool, the failure type, the root cause, and the focus. These carry context only.
    \item \textbf{Contract fields.} Required measurement, required signature, success criterion, scope, anti-scope, matcher, and an optional calibration request. These are checked repeatedly downstream.
    \item \textbf{Global planner.} It merges proposals that ask for the same measurement and narrows scopes that overlap. Merging incompatible contracts is rejected mechanically rather than discouraged in the prompt, and after two failed attempts the round falls back to the validated local plans as singleton groups and continues.
\end{itemize}

  \item \textbf{Synthesis.} Each work group gets at most $K_{\mathrm{update}}{=}3$ sequential attempts.
\begin{itemize}[leftmargin=1.2em, topsep=2pt, itemsep=1pt, parsep=0pt]
    \item \textbf{What the synthesizer sees.} The current library, rendered separately, together with a digest carrying the bucket dossier and its support examples, the planner's diagnosis, the contract, a compact timeline of previous failed attempts, and the numerically strongest previous attempt kept apart so that a weak later retry cannot displace it.
    \item \textbf{Hard constraints.} The prompt fixes a single-function output with no \texttt{while} loops and a zero-argument signature.
    \item \textbf{Checks before execution.} A single-function check, a smoke call built from the signature, and a context-selection check against the contract.
    \item \textbf{Calibration.} When a contract asks for a decision boundary, the tool may carry a shallow decision tree of kind \texttt{binary\_decision\_tree\_v1}, fitted on training rows alone with validation and test never read and answers stored only as hashes. The tree needs at least 16 samples per class and an internal 80/20 selection split grouped by input fingerprint, and it is accepted only at an internal balanced accuracy of at least $\beta = 0.55$. An accepted tree is AST-compiled into the tool source, and a calibrated tool may not call the forecasting primitive.
    \item \textbf{Retry policy.} A contract-violating artifact receives one targeted repair, an undeliverable input signature is adapted between retries, and two consecutive zero-repair outcomes trigger a replan that may change the artifact kind, for example from a new tool to usage guidance.
\end{itemize}

  \item \textbf{Candidate prefilter.} Each candidate is screened cheaply before any candidate reaches the round-level gate.
\begin{itemize}[leftmargin=1.2em, topsep=2pt, itemsep=1pt, parsep=0pt]
    \item \textbf{Sampling.} The first 32 held-out errors of the candidate's own bucket, taken in a deterministic order that interleaves the members of a coordinated work group so that a bounded screen does not sample only the largest source bucket.
    \item \textbf{Attribution.} A newly created tool is first checked for signature deliverability on those questions, while lifecycle updates are attributed through action-specific traces instead, because narrowing or rescoping can help precisely by preventing delivery.
    \item \textbf{Criterion.} At least one attributable repair, meaning a question fixed with the tool bound, executed, and visible in the trace, within coarse caps on harm and execution errors.
    \item \textbf{Stopping.} A repair rate of at least $0.40$ stops the search early, and otherwise the best positive candidate after three attempts still reaches the gate.
\end{itemize}

  \item \textbf{Whole-library gate.} The current library and the candidate library each answer the full validation split, and only questions both arms completed are compared, so helped and harmed are counted on identical questions.
\begin{itemize}[leftmargin=1.2em, topsep=2pt, itemsep=1pt, parsep=0pt]
    \item \textbf{Statistic.} With $n$ paired questions the gate computes $\delta = (h-m)/n$ and $\delta_{\mathrm{lb}} = \delta - 1.96\sqrt{\max(h+m,1)}/n$.
    \item \textbf{Strict path.} A library passes when $h > 0$ and $\delta_{\mathrm{lb}} > 0$.
    \item \textbf{Single-round path.} A single-round run has no later round in which to accumulate evidence, so a supplementary rule admits a library when $\delta > 0$, the net repair is at least $\max(3, \lceil 0.02n \rceil)$, and the absolute harm rate is at most $0.10$.
    \item \textbf{Invalidity.} An execution error rate above $2\%$ renders the measurement invalid, which is separated from a quality failure.
\end{itemize}

  \item \textbf{One-shot attribution pruning.} On a failed gate with clearly harmful updates, all of them and their dependents are deleted in one transaction, the pruned library is re-verified in exactly one more full paired run, and it is deployed only on strict improvement together with a passing gate, else the round rolls back. There is no second prune.

  \item \textbf{Chain write and report-only test.} An accepted library first reconciles its declared scope fields against how its tools actually behaved, and because that edit changes the library it passes the same paired gate one final time. The round then appends exactly one repair stage to the chain. The test split is answered once per round for reporting only, and on a rollback the best failed candidate may receive one diagnostic test run that participates in no selection.
\end{enumerate}

\subsection{Frozen Configuration}
\label{sec:frozenconfig}

\cref{tab:config1} lists every knob that was fixed before the first run and never touched again, so that a reader can reproduce a round without reading the code.
Three groups appear in the table.
The first fixes the models and the shape of a round: all four agent roles run on the same backbone, synthesis runs at a higher reasoning effort because it writes code, the root library is empty, and a run is a single round with at most three synthesis attempts per work group.
The second fixes the two levels of validation: the prefilter samples 32 held-out errors and demands one attributable repair, while the gate compares the whole library on the full validation split at $Z = 1.96$ and admits either through the strict lower bound or through the single-round rule with a net floor of three and a harm ceiling of $0.10$.
The third fixes how the agent answers: it consults a structured scope catalog before inspecting at most three tools, tools return numbers rather than verdicts, and no hand-written prompt is added for any task.
Two entries deserve emphasis, because they are what makes a rejected round harmless: \texttt{AUTO\_PRUNE} is off so that only one bounded prune can run, and \texttt{TEST\_REJECTED\_CANDIDATE} is on so that a rolled-back library still receives one diagnostic test run that participates in no selection.

\begin{table}[h]
\footnotesize
\caption{The frozen protocol. Every reported run uses these settings without exception.}
\label{tab:config1}
\setlength{\tabcolsep}{5pt}
\begin{center}
\resizebox{\textwidth}{!}{%
\begin{tabular}{@{}lll@{}}
\toprule
\textbf{Knob} & \textbf{Value} & \textbf{Meaning} \\
\midrule
\multicolumn{3}{l}{\textit{Models and evolution}} \\
\texttt{TASK\_LM} / \texttt{TRIAGE\_LM} / \texttt{PLANNER\_LM} & luna, effort low & all agent roles \\
\texttt{SYNTH\_LM} & \texttt{luna\_synth}, effort medium & tool synthesis \\
\texttt{ROOT\_LIBRARY} & \texttt{empty} & evolution starts from no tools \\
\texttt{RESIDUAL\_REPAIR} & \texttt{1} & residual chain, mandatory \\
\texttt{TOOL\_CALIBRATION} & \texttt{1} & train-only calibration on \\
\texttt{ROUNDS} / $K$ / \texttt{K\_KIND} / \texttt{K\_UPDATE} & \texttt{1 / 3 / 1 / 3} & single round, three retries \\
\texttt{SUPPORT\_MAX} / \texttt{BACKGROUND\_N} & \texttt{12 / 48} & planner-visible samples \\
\texttt{MAX\_FAILURES} / \texttt{MAX\_BUCKETS} & \texttt{all / all} & no truncation \\
\texttt{ROUND\_PLAN\_COORDINATION} & \texttt{1} & global coordinator on \\
\midrule
\multicolumn{3}{l}{\textit{Prefilter and gate}} \\
\texttt{PREFILTER\_N} / \texttt{MIN\_HELPED} & \texttt{32 / 1} & prefilter sample and floor \\
\texttt{CANDIDATE\_EARLY\_STOP\_REPAIR} & \texttt{0.40} & early-stop threshold \\
\texttt{GATE\_MODE} / \texttt{LLM\_GATE} & \texttt{library / 0} & whole-library gate, no LLM judge \\
\texttt{LIBGATE\_N} / \texttt{LIBGATE\_Z} & \texttt{all} / \texttt{1.96} & full split, lower-bound $Z$ \\
\texttt{ONE\_ROUND\_SAFE\_POSITIVE} & \texttt{1} & safe-positive rule active \\
\texttt{SAFE\_MIN\_NET} / \texttt{SAFE\_MAX\_HARM\_RATE} & \texttt{3 / 0.10} & net floor, harm ceiling \\
\texttt{AUTO\_PRUNE} / \texttt{ONE\_SHOT\_ATTRIBUTION\_PRUNE} & \texttt{0 / 1} & one bounded prune only \\
\texttt{RESCOPE\_AFTER\_ROUND} & \texttt{1} & scope recalibration before chain write \\
\texttt{TEST\_REJECTED\_CANDIDATE} & \texttt{1} & diagnostic test on rollback \\
\midrule
\multicolumn{3}{l}{\textit{Routing and answering}} \\
\texttt{ROUTING} / \texttt{ENCODING} / \texttt{CONTEXT} & \texttt{inline / shape / preview} & scope catalog first, $\leq 3$ inspects \\
\texttt{TOOL\_VERDICT} / \texttt{STRONG\_PROMPT} / \texttt{TS\_PROMPT} & \texttt{0 / 0 / 0} & no verdicts, no hand-written prompts \\
\texttt{BASE\_CACHE} / \texttt{EVAL\_CACHE} & \texttt{1 / 0} & base answers cached once per run \\
\bottomrule
\end{tabular}}
\end{center}
\end{table}

\section{Full Ablation Results}
\label{sec:supp}

\subsection{Component Ablations}
\label{sec:supp_components}

\cref{tab:ablation} is the numeric table behind~\cref{fig:ablation}a, with one row per configuration on each of the four cells.
The two \emph{Val} columns are what the gate actually saw before it made its decision: the paired delta on the validation split, its lower bound, and the helped and harmed counts behind them.
Reading them next to \emph{Final} shows how well the gate's own evidence predicted the deployed outcome, which is the calibration plotted in~\cref{fig:cards}b.
The \emph{Rejected diagnostic} column is filled only for rounds the gate rejected. Such a round deploys nothing, so its final accuracy equals its base, and the diagnostic reports what the rejected library would have scored had it been deployed, measured once and used in no decision.
\emph{Final} carries the gain the round itself produced, \emph{Gap to full} compares it against the full method on the same cell, and \emph{Tools} gives the size of the deployed library, with \emph{rejected} marking a round that shipped none.

\begin{table}[t]
\scriptsize
\caption{Component ablations on TimeMQA Anomaly~(main) and TSAQA-DT~(reproduction), each under two backbones. Val columns show what the gate saw~(paired delta, lower bound, helped/harmed). Base and Final are that run's own test accuracy before and after the round, with the gain the round itself produced in parentheses, and the last column gives the gap to the full method. A rejected library is not deployed, so its final accuracy equals its base and the diagnostic column reports what the rejected library would have scored. Calibration is marked N/A on TSAQA-DT, whose libraries carry no calibration trees.}
\label{tab:ablation}
\setlength{\tabcolsep}{3pt}
\begin{center}
\resizebox{\textwidth}{!}{%
\begin{tabular}{llrrrrrr}
\toprule
Dataset / backbone & Configuration & Val $\delta$ / $\delta_{\mathrm{lb}}$~(h/m) & Rejected diagnostic & Base & \cellcolor{secondcell}Final~(evo gain) & Gap to full & Tools \\
\midrule
\rowcolor{bandA}Anomaly / mini & full method & $+12.25$ / $+7.63$~(69/20) & --- & $58.75$ & $\mathbf{75.00}$~{\tiny($+16.25$)} & --- & 7 \\
 & w/o calibration & $-0.50$ / $-1.70$~(2/4) & $-0.75$ & $63.50$ & \cellcolor{secondcell}$63.50$~{\tiny($+0.00$)} & $-16.25$ & 0~(rejected) \\
 & w/o gate & $-2.00$ / $-5.67$~(24/32) & --- & $63.00$ & \cellcolor{secondcell}$61.75$~{\tiny($-1.25$)} & $-17.50$ & 4 \\
 & w/o scope & $+5.25$ / $+1.06$~(47/26) & --- & $63.00$ & \cellcolor{secondcell}$70.75$~{\tiny($+7.75$)} & $-8.50$ & 5 \\
 & w/ chronos root & $-0.25$ / $-2.15$~(7/8) & $+3.25$ & $62.00$ & \cellcolor{secondcell}$62.00$~{\tiny($+0.00$)} & $-16.25$ & 1 \\
 & w/ 21 tools root & $+0.50$ / $-2.00$~(14/12) & $-1.25$ & $63.00$ & \cellcolor{secondcell}$63.00$~{\tiny($+0.00$)} & $-16.25$ & 21 \\
\midrule
\rowcolor{bandA}Anomaly / terra & full method & $+13.50$ / $+9.06$~(68/14) & --- & $62.50$ & $\mathbf{75.00}$~{\tiny($+12.50$)} & --- & 8 \\
 & w/o calibration & $-4.50$ / $-9.50$~(43/61) & $-4.25$ & $62.50$ & \cellcolor{secondcell}$62.50$~{\tiny($+0.00$)} & $-12.50$ & 0~(rejected) \\
 & w/o gate & $+3.50$ / $-0.83$~(46/32) & --- & $57.50$ & \cellcolor{secondcell}$63.75$~{\tiny($+6.25$)} & $-6.25$ & 8 \\
 & w/o scope & $+10.50$ / $+5.96$~(64/22) & --- & $58.50$ & \cellcolor{secondcell}$70.75$~{\tiny($+12.25$)} & $-0.25$ & 8 \\
 & w/ chronos root & $+16.25$ / $+10.90$~(92/27) & --- & $56.75$ & \cellcolor{secondcell}$71.75$~{\tiny($+15.00$)} & $+2.50$ & 9 \\
 & w/ 21 tools root & $+10.50$ / $+6.28$~(58/16) & --- & $53.75$ & \cellcolor{secondcell}$64.25$~{\tiny($+10.50$)} & $-2.00$ & 26 \\
\midrule
\rowcolor{bandA}TSAQA-DT / mini & full method & $+19.75$ / $+15.39$~(79/0) & --- & $67.75$ & $\mathbf{83.75}$~{\tiny($+16.00$)} & --- & 4 \\
 & w/o calibration & N/A & N/A & N/A & \cellcolor{secondcell}N/A & N/A & N/A \\
 & w/o gate & $+0.50$ / $-1.33$~(8/6) & --- & $65.50$ & \cellcolor{secondcell}$68.75$~{\tiny($+3.25$)} & $-12.75$ & 4 \\
 & w/o scope & $+6.25$ / $+3.52$~(28/3) & --- & $68.75$ & \cellcolor{secondcell}$74.50$~{\tiny($+5.75$)} & $-10.25$ & 6 \\
 & w/ chronos root & $+3.00$ / $+1.17$~(13/1) & --- & $61.00$ & \cellcolor{secondcell}$65.25$~{\tiny($+4.25$)} & $-11.75$ & 2 \\
 & w/ 21 tools root & $+0.50$ / $-1.05$~(6/4) & $+1.00$ & $62.25$ & \cellcolor{secondcell}$62.25$~{\tiny($+0.00$)} & $-16.00$ & 21 \\
\midrule
\rowcolor{bandA}TSAQA-DT / terra & full method & $+15.00$ / $+10.67$~(69/9) & --- & $71.00$ & $\mathbf{85.50}$~{\tiny($+14.50$)} & --- & 4 \\
 & w/o calibration & N/A & N/A & N/A & \cellcolor{secondcell}N/A & N/A & N/A \\
 & w/o gate & $+15.50$ / $+11.64$~(62/0) & --- & $72.43$ & \cellcolor{secondcell}$84.71$~{\tiny($+12.28$)} & $-2.22$ & 4 \\
 & w/o scope & $+14.50$ / $+10.64$~(60/2) & --- & $71.50$ & \cellcolor{secondcell}$84.50$~{\tiny($+13.00$)} & $-1.50$ & 2 \\
 & w/ chronos root & $+11.50$ / $+7.23$~(61/15) & --- & $69.42$ & \cellcolor{secondcell}$82.46$~{\tiny($+13.04$)} & $-1.46$ & 3 \\
 & w/ 21 tools root & $+3.25$ / $+0.52$~(22/9) & --- & $77.75$ & \cellcolor{secondcell}$77.25$~{\tiny($-0.50$)} & $-15.00$ & 25 \\
\bottomrule
\end{tabular}}
\end{center}
\end{table}

\subsection{Residual Chain against Overwrite}
\label{sec:supp_residual}

The residual chain is not a removed component but a second way of applying the same library, so~\cref{tab:residual} reports it separately.
Both arms deploy the same tools on the same questions, and the only difference is whether an admitted tool reviews the base answer or replaces it outright.
The two arms are what~\cref{fig:ablation}b draws.

\begin{table}[t]
\footnotesize
\caption{The residual chain against unconditional overwrite. Both arms deploy the same library on the same questions, and the only difference is whether the library reviews the base answer or replaces it outright. Fixed counts previously wrong answers the update repaired, and harmed counts previously correct answers it broke.}
\label{tab:residual}
\setlength{\tabcolsep}{7pt}
\begin{center}
\begin{tabular}{@{}lrrrr@{}}
\toprule
& \multicolumn{2}{c}{\textbf{Fixed}} & \multicolumn{2}{c}{\textbf{Harmed}} \\
\cmidrule(lr){2-3}\cmidrule(lr){4-5}
\textbf{Dataset / backbone} & Chain & Overwrite & Chain & Overwrite \\
\midrule
Anomaly / mini    & 33 & 58 & 19 & 59 \\
Anomaly / terra   & 77 & 93 & 22 & 32 \\
TSAQA-DT / mini   & 56 & 77 & 3  & 26 \\
TSAQA-DT / terra  & 59 & 78 & 10 & 28 \\
\bottomrule
\end{tabular}
\end{center}
\end{table}

\subsection{Evolution Funnel across Tasks and Backbones}
\label{sec:supp_funnel}

\cref{tab:funnel_summary} summarizes attrition across the one-round process, and \cref{tab:funnel_tasks} opens the full trace of one complete GPT-5.6-luna run per task, one row per task.
Every field is read from that run's sole round, with no cross-run imputation and no missing fields.

Across the 30 runs, 4{,}898 of 4{,}910 errors were assigned to a bucket: only 12 remained unassigned and none were discarded for an undersized bucket.
Coordination reduced 197 local proposals to 194 contracts, while nine of 203 failure buckets, spread across seven runs, produced no candidate.
The prefilter retained 227 of 483 synthesis attempts~(47.0\%), after which 113 artifacts were deployed: 108 tools, five prompt patches, and no usage-only artifacts.
The recorded final outcomes were 22 passes, seven passes after one-shot pruning~(17 artifacts removed in total), and one rollback.
Thus most cost occurs at executable synthesis, the prefilter, and the library gate rather than during clustering or coordination.

Funnel width reflects error diversity rather than the eventual payoff.
For example, the T3 run has 367 errors, 11 buckets, 31 synthesis attempts, and six deployed artifacts, yet a test gain of $+7.43$~pp, while the MMTS Match run has only 28 errors, three buckets, six attempts, and two deployed artifacts, yet gains $+18.75$~pp.
More failures therefore create more search work, but not necessarily more useful headroom.

\begin{table}[h!]
\small
\caption{Aggregate one-round evolution funnel over the 30 task--backbone runs. Totals count events across runs, and medians and ranges are computed per run. Yield compares each row with the stage above it.}
\label{tab:funnel_summary}
\setlength{\tabcolsep}{7pt}
\begin{center}
\begin{tabular}{@{}lrrr@{}}
\toprule
\textbf{Stage} & \textbf{Total} & \textbf{Median [min--max]} & \textbf{Yield} \\
\midrule
Errors on the training split  & 4{,}910 & 156 [26--369]   & --- \\
Errors assigned to a bucket   & 4{,}898 & 155.5 [24--369] & 99.76\% \\
Buckets                       & 203     & 6 [2--12]       & --- \\
Buckets that produced a candidate & 194 & 6 [2--12]       & 95.57\% \\
Local proposals               & 197     & 6 [2--12]       & --- \\
Contracts after coordination  & 194     & 6 [2--12]       & 98.48\% \\
Synthesis attempts            & 483     & 16 [4--32]      & --- \\
Candidates past the prefilter & 227     & 7 [1--17]       & 47.00\% \\
Deployed artifacts            & 113     & 3.5 [0--8]      & 49.78\% \\
\bottomrule
\end{tabular}
\end{center}
\end{table}

\begin{table}[t]
\scriptsize
\caption{One-round evolution funnel of one complete GPT-5.6-luna run per task, recorded end to end for process analysis. \emph{Errors} is the base pass's error count on the training split and \emph{Tagged} the errors assigned to a bucket. \emph{Local} and \emph{Contracts} are proposals before and after coordination, \emph{Synth.} counts every synthesis attempt, \emph{Prefilter} the attempts that passed it, and \emph{Deploy T/P} the deployed tools and prompt patches. \emph{Final gate} records a direct pass, a rollback, or the number of artifacts removed by the one-shot prune before a passing re-verification. Each test delta is that run's own paired delta rather than the number reported in~\cref{tab:main}.}
\label{tab:funnel_tasks}
\setlength{\tabcolsep}{2pt}
\renewcommand{\arraystretch}{1.15}
\begin{center}
\resizebox{\textwidth}{!}{%
\begin{tabular}{l|cccccccccc}
\toprule
\textbf{Task} & \textbf{Errors} & \textbf{Tagged} & \textbf{Buckets} & \textbf{Local} & \textbf{Contracts} & \textbf{Synth.} & \textbf{Prefilter} & \textbf{Deploy T/P} & \textbf{Final gate} & $\boldsymbol{\Delta}_{\mathrm{test}}$ \textbf{(pp)} \\
\midrule
\rowcolor{bandA}T1 & 246 & 246 & 8 & 8 & 8 & 20 & 11 & 4/0 & prune 3 $\rightarrow$ pass & $+9.49$ \\
\rowcolor{bandB}T2 & 218 & 218 & 8 & 8 & 8 & 22 & 8 & 4/0 & pass & $+12.70$ \\
\rowcolor{bandC}T3 & 367 & 367 & 11 & 11 & 11 & 31 & 12 & 6/0 & pass & $+7.43$ \\
\rowcolor{bandD}T4 & 198 & 198 & 5 & 5 & 4 & 11 & 4 & 2/0 & pass & $+12.70$ \\
\rowcolor{bandE}TSExam & 119 & 119 & 8 & 8 & 8 & 22 & 13 & 6/0 & pass & $+8.72$ \\
\rowcolor{bandF}Merrill & 27 & 25 & 6 & 4 & 4 & 10 & 7 & 2/1 & pass & $+2.50$ \\
\rowcolor{bandG}Anomaly & 164 & 164 & 8 & 8 & 8 & 12 & 11 & 8/0 & pass & $+19.75$ \\
\rowcolor{bandH}Classification & 158 & 158 & 6 & 6 & 6 & 16 & 1 & 1/0 & pass & $+26.25$ \\
\rowcolor{bandI}Match & 28 & 28 & 3 & 3 & 3 & 6 & 4 & 2/0 & prune 1 $\rightarrow$ pass & $+18.75$ \\
\rowcolor{bandJ}TSAQA-DT & 121 & 121 & 7 & 7 & 6 & 14 & 4 & 3/0 & pass & $+13.75$ \\
\bottomrule
\end{tabular}}
\end{center}
\end{table}

\end{document}